\documentclass{opt2026}

\usepackage{booktabs}       

\usepackage[framemethod=tikz]{mdframed}
\usepackage{adjustbox}
\usepackage{mathtools}
\usepackage{float}
\usepackage{multicol}
\usepackage{xcolor}
\usepackage{enumitem}
\usepackage[autostyle]{csquotes}
\usepackage[none]{hyphenat}
\RestyleAlgo{ruled}

\newtheorem{assumption}{Assumption}
\newcommand{\subtitleblock}[2]{%
    \textbf{\small \textit{#1}}
    
    {\leftskip=4em
    #2\par
    }
}

\title[BOND]{BOND: License to Train with Black-Box Functions}

\optauthor{%
    \Name{Andrew Clark} \Email{andrew.clark2@cba.com.au}\\
    \Name{Osmaan Rasouli}\\
    \Name{Jack Moursounidis}\\
    \Name{William Gan}\\
    \Name{Cooper Doyle}\\
    \addr Commonwealth Bank of Australia
}

\begin{document}

\maketitle

\begin{abstract}
    We introduce Bounded Numerical Differentiation (BOND), a perturbative method for estimating the gradients of black-box functions. BOND is distinguished by its formulation, which adaptively bounds perturbations for reliable gradient signs, and by its implementation, which operates at black-box interfaces. This enables BOND to be more accurate and scalable compared to existing methods, facilitating end-to-end training of architectures that incorporate non-autodifferentiable modules. We observe that these modules, implemented in our experiments as frozen networks, can enhance model performance without increasing trainable parameter count. Our findings highlight the potential of leveraging fixed transformations to expand model capacity, pointing to hybrid analogue–digital devices as a path to scaling networks, and provides insights into the dynamics of adaptive optimizers.
\end{abstract}

\section{Introduction}
Representing high-dimensional functions using trainable neural network architectures is effective, but computationally demanding. 
The notion that these functions could, in part, be efficiently replaced or augmented with black-box transforms on specialized hardware is a core motivation for our work. These components are \enquote{black-box} in the sense that they lack a closed-form expression and do not explicitly maintain or propagate a computational graph, precluding their compatibility with conventional Auto-Differentiation (AD). 

Without a computational graph, black-box functions can still be integrated in network architectures using gradient estimation. Zeroth-Order (ZO) methods estimate gradients from perturbed function evaluations, enabling optimization without access to exact derivative information \citep{chen2017zoo,ilyas2018black,liu2020primer}, though many such methods have been limited by the trade-off between estimation complexity and accuracy. 

In this paper, we introduce Bounded Numerical Differentiation (BOND), a new ZO variant which demonstrates similar convergence rates to First-Order (FO) counterparts, with reduced estimation complexity and hyperparameter dependencies compared to Finite Difference Stochastic Approximation (FDSA) \citep{KieferWolfowitz1952Stochastic,fdsa_convergence}. This is achieved using adaptive perturbation bounds and estimating derivatives only at black-box interfaces, enabling backpropagation across the local computational graphs of surrounding white-box networks. BOND is technically compatible with any black-box function that admits a forward pass, but it relies on standard assumptions such as smoothness for optimal performance. This enables training a \textit{read-in} network upstream of the black-box function, extending the conventional Reservoir-Computing (RC) architecture, which typically trains only the \textit{read-out} network. As our work lies at the intersection of RC and ZO methods, we use the terms \enquote{reservoir} and \enquote{black-box function} interchangeably throughout.

Our main contributions are: 1) We introduce Bounded Numerical Differentiation (BOND), a new ZO variant which enables accurate gradient estimations across functions which do not permit autograd calculations. 2) We use BOND to explore a basic framework which integrates black-box functions in a trainable network architecture, equivalent to training a \textit{read-in} network in an RC \citep{verstraeten2007experimental,336133} context.

\section{Related Work} \label{section:related_work}
\subsection{Zeroth-Order Methods}

ZO methods use perturbed function evaluations to obtain numerical derivative estimates, rather than explicit gradient calculation \citep{liu2020primer,robbinsmonro1951stochastic,flaxman2004bandit}. Among ZO variants, a key distinction is made between approaches using FDSA \citep{Bhatnagar2013}, and SPSA \citep{OGSPSASpall, SpallSPSAonemeasurement, SPSA2001SpallOverview}. As parameter dimension $d_{\theta}=|\theta|$ increases, SPSA variants search relatively fewer directions compared to FDSA, making them more efficient at the cost of increased gradient noise, which typically leads to a reduction in performance \citep{signSGD_liu,liu_variance_reduction}. Existing theory shows that SPSA convergence can approach FO methods, where the gap increases as parameter dimension $d_{\theta}$ increases \citep{zo_adam,duchi_convergence}. Increasing the perturbation count $n_{\textit{pert}}$ can also improve convergence but at the cost of more function evaluations, which in the limit is equivalent to a coordinate-wise approach \citep{duchi_convergence}. Notably, these methods require a suitable choice of smoothing coefficient $\mu_t$, the selection of which becomes non\-trivial in the presence of numerical or measurement noise, i.e., in the application of a physical device. In this setting, BOND seeks to both reduce the scaling limitation of $\mathcal{O}(d_{\theta})$ estimation complexity, and alleviate the hyperparameter dependency of $\mu_t$.

\subsection{Black-Box and Reservoir Architectures}

Reservoir Computing (RC) provides a uniquely-suited setting for incorporating black-box transformations into trainable architectures. Previous research exploring ZO methods in black-box architectures yielded promising results, albeit not as successful as AD \citep{multiplex}, or remained limited by SPSA-type error as in \cite{pgt}. By enabling the training of \textit{read-in} networks, BOND allows for new explorations in the RC setting. A deeper analysis of the literature on black-box architectures is provided in Appendix~\ref{appendix:related_works}.

\section{Method} \label{method}
\subsection{Overview}
\begin{figure}[H]
    \centering
    \includegraphics[width=0.5\textwidth]{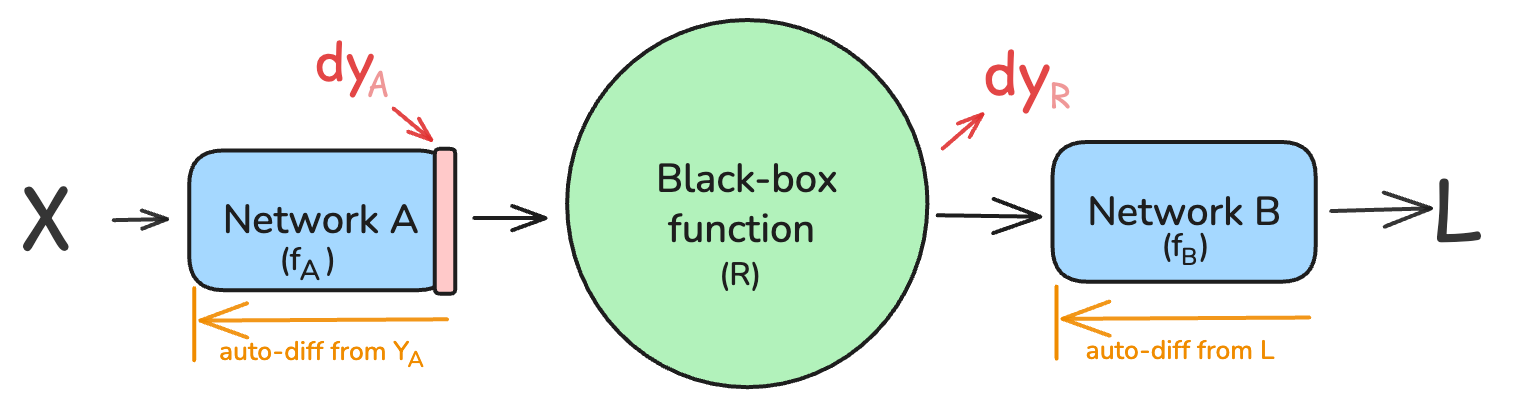}
    \caption{Schematic of reservoir $\mathcal{R}(\cdot)$ with trainable \textit{read-in} and \textit{read-out} networks.}
    \label{fig:NFN_diagram}
\end{figure}

In conventional RC setups, particularly those using physical devices, linear transforms are often used to map data to the input dimensions of a reservoir $\mathcal{R}(\cdot)$ \citep{Gauthier_2021,lukovsevivcius2012reservoir,verstraeten2007experimental}. This approach, however, is likely to suffer from suboptimal utilization of the reservoir functionality, particularly as data dimensionality increases. To maximize performance, we require an optimized mapping $f_A(X) \mapsto Y_{A}$ from data $X$ to the reservoir inputs $Y_{A}$, and reservoir outputs $Y_{\mathcal{R}}$ to the target variable $f_B(Y_{\mathcal{R}}) \mapsto Y_B$, where $L = \mathcal{L}(Y,Y_B)$. This requires a trainable \textit{read-in} ($f_A$) and  \textit{read-out} ($f_B$) network, illustrated schematically in Figure \ref{fig:NFN_diagram}. 
Without a computational graph across $\mathcal{R}$, the gradients in $f_A$ must be estimated. However, by estimating the partials only at the reservoir interface $\hat{\frac{\partial Y_{\mathcal{R}}}{\partial Y_{A}}}$, we can implement AD across a cached $f_A$ computational graph. This follows from the chain rule: substituting the estimated interface partial $\hat{\frac{\partial Y_{\mathcal{R}}}{\partial Y_{A}}}$ for the true (inaccessible) partial $\frac{\partial Y_{\mathcal{R}}}{\partial Y_{A}}$.
By estimating $\hat{\frac{\partial Y_{\mathcal{R}}}{\partial Y_{A}}}$ then using AD, we only require as many function evaluations as there are inputs to the reservoir $d_{\mathcal{R}}$. This reduces the FDSA estimation complexity from $\mathcal{O}(d_{\theta_A})$ to $\mathcal{O}(d_{\mathcal{R}})$, where $d_{\mathcal{R}} \ll d_{\theta_A}$. Through this mechanism, BOND does not impose any limit on $d_{\theta_{A}}$, enabling the use of large network architectures. In practice, especially for physical devices, we acknowledge that the requirement of $d_{\mathcal{R}}$ function evaluations is still not entirely optimal. 

\subsection{Implementation}
Inspired by sign-based optimization \citep{signSGD_liu,liu2020primer}, BOND's perturbation bounds are designed for the accurate estimation of gradient sign, accounting for the degradation of smoothness as perturbation magnitude goes to $0$, and the increasing uncertainty in sign direction associated with larger perturbations. The detailed workings underlying the BOND pertubation bounds are provided in Appendix~\ref{appendix:proofs}, with this section containing main observations and limitations. The formulations for the upper $\overline{\Delta_x^t}$ and lower $\underline{\Delta_x^t}$ bounds are shown below, with $\epsilon=1e-8$ for numerical stability.
\begin{equation}
    \underline{\Delta_x^t} = \frac{b}{\min \big\{|\hat{m}^{\mathcal{R}}_{t-1}|\big\} + \epsilon}\label{eq:lower_bound_used}
\end{equation}
\begin{equation}
\overline{\Delta_x^t} = \alpha_{t} \cdot \sqrt{\frac{|\hat{m}^{\mathcal{R}}_{t-1}|}{\hat{v}^{\mathcal{R}}_{t-1} + \epsilon}}\label{eq:upper_bound_used}
\end{equation}
The BOND algorithm (Algorithm~\ref{pseudo_code_bond}) uniformly samples perturbations $\Delta^t_x \in \mathbb{R}^{|Y_A|}$ from the bounds in Equations~\ref{eq:lower_bound_used} and \ref{eq:upper_bound_used} to compute input-wise central-difference gradient estimates at the black-box interface. We also use a variant with simultaneous perturbations (BONDS), which is detailed with supporting algorithms in Appendix~\ref{appendix:algorithms}. We observe there is meaningful similarity between our upper bound formulation in Equation~\ref{eq:upper_bound_used} and the Adam parameter-update formula~\citep{adam}. As opposed to Adam, which sets $\alpha_{t}=\eta_{t}$ where $\eta_t$ is the learning rate, we use $\alpha_{t}=\max{|\hat{m}^{\mathcal{R}}_{t-1}|}$, which yields reasonable empirical results and reduces reliance on hyperparameters. Equation \ref{eq:upper_bound_used} also creates a unique upper bound for each input dimension, which is less restrictive than a global smoothing coefficient as in SPSA. Note, BOND assumes a three-times-differentiable $\mathcal{R}$ (Assumption~\ref{assump:diff_3}), which is not uncommon in existing works \citep{scaling_rnns,stoch_approx_high_dimensions,fdsa_convergence}, though this renders BOND's success contingent on the smoothness properties of $\mathcal{R}$, motivating our implementation of Tanh activations. BOND also assumes uniform baseline noise
(Assumption~\ref{assumption:uniform_noise}). Details are discussed further in Appendix~\ref{appendix:proofs} and \ref{appendix:perturbation_magnitudes}. 


\section{Experiments} \label{section-experiments}
In this section, we present experiments based on the NEN, NFN(\textit{Fixed}) and NFN(\textit{Parallel}) configurations detailed in Appendix~\ref{appendix:reservoir_functions}. For clarity, these configurations are based on the Network-Reservoir-Network architecture in Figure~\ref{fig:NFN_diagram}, where N\textbf{E}N and N\textbf{F}N refer respectively to the use of a frozen \textit{\textbf{E}choStateNetwork} and \textit{\textbf{F}eedFordwardNueralNetwork} (FFNN) as the reservoir. \textit{Fixed} and \textit{Parallel} denote fully connected and partitioned FFNN configurations, see Appendix~\ref{appendix:reservoir_functions}. We also introduce an implementation with a pre-trained reservoir, NFN(\textit{Rockmelon}), which is directly modelled on the input-output pairs of a physical RC device, see Appendix~\ref{appendix:setup_rockmelon}. Unlike the random \textit{Fixed} and \textit{Parallel} functions, \textit{Rockmelon} exhibits structured correlations approximately representative of a specialized hardware system.

\subsection{Sign Accuracy and Convergence}
\begin{figure}[ht]
    \centering
    \includegraphics[width=0.8\textwidth]{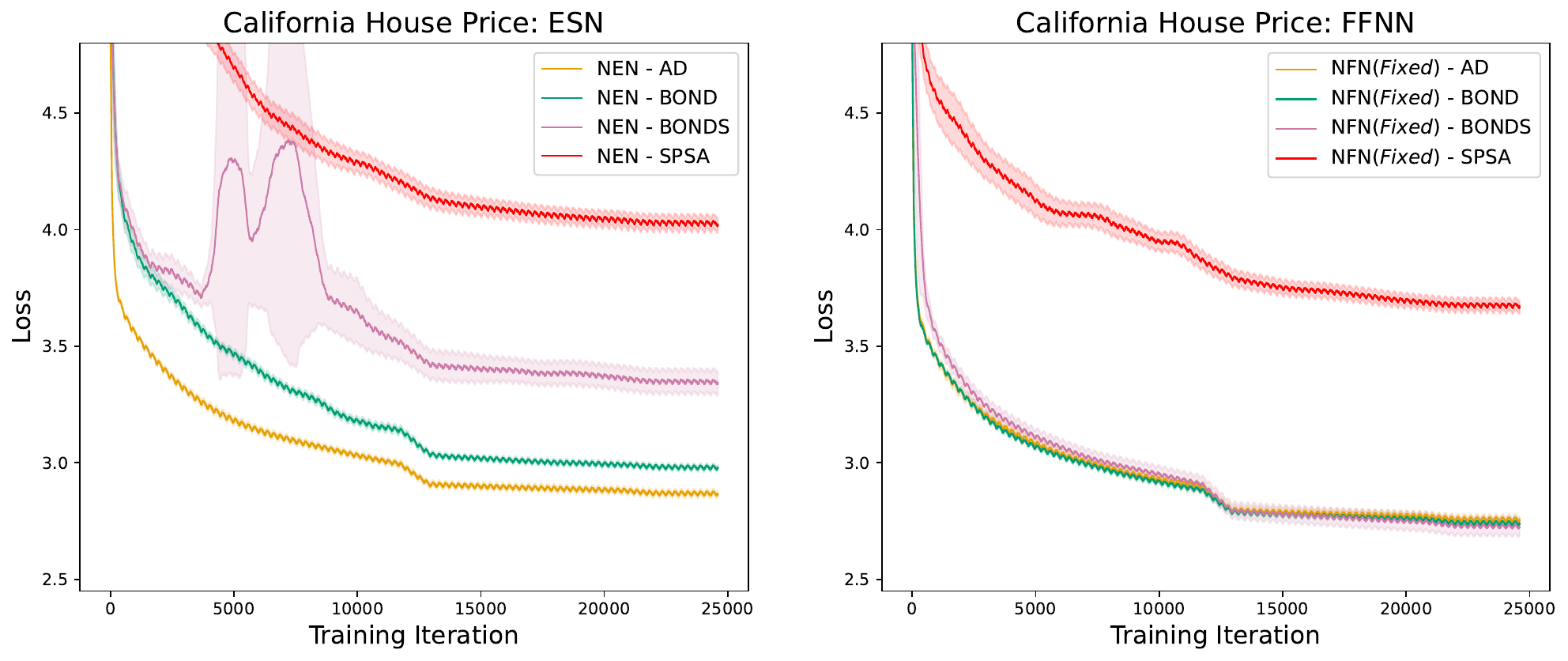}
    \caption{Comparing gradient estimation methods using the NEN and NFN(\textit{Fixed}) model architectures on the California House Price Dataset.}
    \label{fig:calhouse_loss}
\end{figure}

Using the California House Price (CHP) regression dataset, we show that, for a frozen black-box function, BOND's gradient estimates yield performance comparable to that of AD, see Figure~\ref{fig:calhouse_loss}. Although CHP data lacks sequential dependencies, NEN serves to demonstrate BOND's behavior on recursive functions and provides context for future time-series applications. We note that each reservoir has 5 inputs and outputs, with full experimental details in Appendix~\ref{appendix:setup_sign_accuracy_and_convergence}.
While BONDS unexpectedly matches AD's convergence for NFN(\textit{Fixed}), we anticipate that higher reservoir input dimensionality will degrade performance by increasing gradient estimation variance. There is a consistent correlation between sign accuracy and performance in the parallel decline from NFN to NEN: BOND decreases $99.95\% \rightarrow 79.86\%$ and BONDS $76.13\% \rightarrow 60.81\%$, see Table~\ref{table:calhouse}, Appendix~\ref{appendix:chp_results}. 
\begin{table}[ht]
    \caption{Comparing gradient estimation methods for NFN(\textit{Fixed}) using ResNet-FashionMNIST and DenseNet-CIFAR10 as \textit{read-in} ($d_{\mathcal{R}}=500$) networks.}
    \label{table:fmnist_cifar10}
    \centering
    \begin{adjustbox}{width=\textwidth}
    \begin{tabular}{lrrrrrr}
        \toprule
        & \multicolumn{3}{c}{FashionMNIST} & \multicolumn{3}{c}{CIFAR10} \\
        \cmidrule(lr){2-4} \cmidrule(lr){5-7}
        Model & \shortstack{Acc. \\ (\%)} & \shortstack{Sign Acc. \\ (\%)} & \shortstack{Loop Time \\ (ms)} & \shortstack{Acc. \\ (\%)} & \shortstack{Sign Acc. \\ (\%)} & \shortstack{Loop Time \\ (ms)} \\
        \midrule
        AD & $93.35_{\pm \scriptstyle 0.21}$ & $-$ & $\textbf{56.18}_{\pm \scriptstyle 0.07}$ & $92.94_{\pm \scriptstyle 0.22}$ & $-$ & $\textbf{254.91}_{\pm \scriptstyle 0.02}$ \\
        BOND & $\textbf{93.42}_{\pm \scriptstyle 0.10}$ & $\textbf{99.91}_{\pm \scriptstyle 0.01}$ & $73.57_{\pm \scriptstyle 0.24}$ & $\textbf{93.25}_{\pm \scriptstyle 0.18}$ & $\textbf{99.99}_{\pm \scriptstyle 0.01}$ & $279.51_{\pm \scriptstyle 0.09}$ \\
        BONDS & $83.25_{\pm \scriptstyle 3.89}$ & $52.07_{\pm \scriptstyle 0.17}$ & $73.58_{\pm \scriptstyle 0.25}$ & $82.95_{\pm \scriptstyle 0.44}$ & $50.54_{\pm \scriptstyle 0.05}$ & $279.32_{\pm \scriptstyle 0.02}$ \\
        SPSA & $26.82_{\pm \scriptstyle 1.61}$ & $50.00_{\pm \scriptstyle 0.01}$ & $108.38_{\pm \scriptstyle 0.56}$ & $42.06_{\pm \scriptstyle 0.38}$ & $50.01_{\pm \scriptstyle 0.01}$ & $452.25_{\pm \scriptstyle 0.01}$ \\
        \bottomrule
    \end{tabular}
    \end{adjustbox}
\end{table}

While BOND incurs computational overhead relative to AD ($\times4.5$ and $\times0.65$ slower for NEN and NFN(\textit{Fixed}) on CHP; Table~\ref{table:calhouse}), its relative constraint decreases as surrounding network capacity $d_{\theta_A}$ grows, as observed on FashionMNIST and CIFAR10 (Table~\ref{table:fmnist_cifar10}) despite scaling $d_{\mathcal{R}}$ from $5$ to $500$. This demonstrates that overhead is primarily dictated by black-box architecture rather than input dimensionality alone, suggesting manageable constraints if device runtime is minimized. SPSA and BONDS performance scales poorly to the CNN \textit{read-in} networks, Table~\ref{table:fmnist_cifar10}, and we note standard FDSA for Figure~\ref{fig:calhouse_loss} would require $23,000$ forward passes per iteration on CHP compared to only $10$ for BOND. Beyond its primary application, BOND offers a framework for probing optimization dynamics; where replacing our upper bound with Adam's parameter-update formula yields slightly lower sign accuracy and performance (Appendix~\ref{appendix:perturbation_magnitudes}, Table~\ref{table:bondvsadam_sign}). Collectively, these results support the following conclusions: BOND accurately estimates gradient sign, achieves comparable convergence to AD for non-recursive functions, and can scale reasonably despite some computational overhead. This uniquely positions BOND among existing ZO methods, enabling explorations on the utility of including black box functions in trainable architectures.

\subsection{Reservoir Scaling}
We use BOND to investigate whether incorporating a black-box function \textit{can} improve model performance. For the CHP dataset, we scale the input dimensionality ($d_{\mathcal{R}}=500$) and the \textit{read-in}/\textit{read-out} networks, comparing against an AD baseline without a reservoir, see Figure~\ref{fig:reservoir_scaling}(a), with experimental details in Appendix~\ref{appendix:setup_calhouse_scaling}. We note that all networks have an equal number of trainable parameters, except NFN(\textit{Rockmelon}), which has fewer to accommodate its 10-input, 2-output sub-reservoir configuration. In Figure~\ref{fig:reservoir_scaling}(a), both black-box functions yield improvements over the \enquote{No Reservoir} baseline when trained with BOND. SPSA and BONDS both hinder convergence, prohibiting the exploration of black-box utility. We find that the pre-trained \textit{Rockmelon} reservoir outperforms the untrained \textit{Parallel} and \textit{Fixed} modules.

\begin{figure}[ht]
    \centering
    \subfigure[California House Price.]{\includegraphics[width=0.48\textwidth]{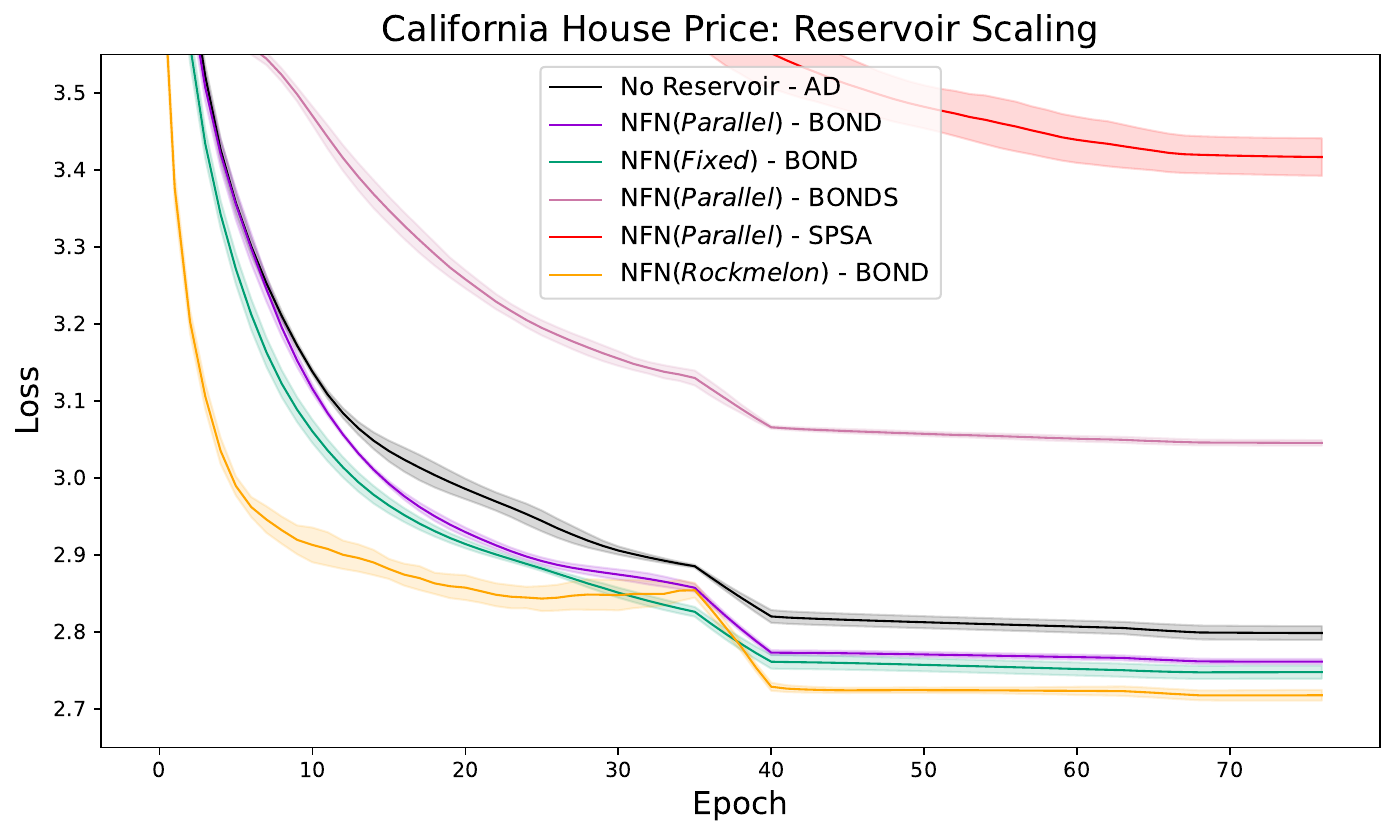}}
    \hfill
    \subfigure[CIFAR-100.]{\includegraphics[width=0.48\textwidth]{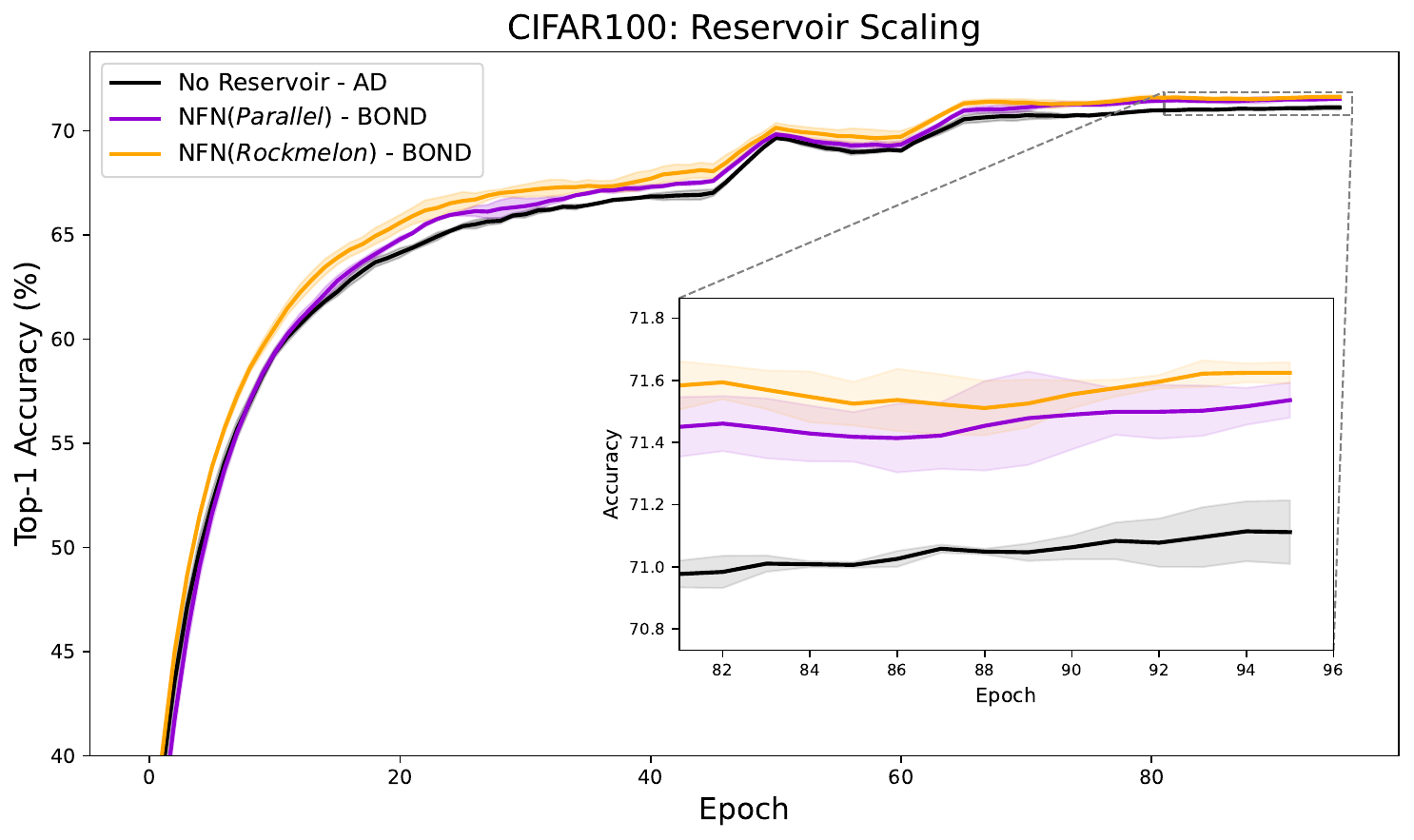}}\label{fig:cifar100}
    \caption{Comparing reservoir-embedded networks trained with ZO methods against AD-trained networks without reservoirs.}
    \label{fig:reservoir_scaling}
\end{figure}
We evaluate a reservoir-embedded model on CIFAR-100 using DenseNet~\citep{densenet} as the \textit{read-in} network, see Figure~\ref{fig:reservoir_scaling}(b). The \textit{read-out} is a compact two-hidden-layer FFNN --- details in Appendix~\ref{appendix:setup_cifar100}. Figure~\ref{fig:reservoir_scaling}(b) demonstrates that NFN(\textit{Parallel}) and NFN(\textit{Rockmelon}) outperform the \enquote{No Reservoir} counterpart by $+0.49\%$ and $+0.53\%$ respectively (full numerical breakdown in Table~\ref{table:cifar100}, Appendix~\ref{appendix:reservoir_scaling}).

\section{Limitations}\label{section-limitations}
BOND incurs computational overheads which make it considerably slower than AD, being only applicable where AD is not possible. To mitigate overhead, we propose a modularized \textit{Parallel} reservoir configuration (Appendix~\ref{appendix:reservoir_functions}) that distributes gradient estimation across multiple sub-reservoirs, enabling parallelism in BOND's perturbations. Even so, scaling BOND to very large reservoir input dimensions $d_{\mathcal{R}}$ may become prohibitive. In this work, we demonstrate that both untrained and trained black-box functions \textit{can} improve performance, but do not characterize precisely what makes reservoirs beneficial. Despite remaining a core motivation, the integration of specialized hardware as black-box functions presents distinct challenges that warrant additional investigation, see Appendix~\ref{appendix:setup_rockmelon}. We also do not provide a theoretical proof or convergence guarantee with BOND, only illustrating support used to formulate the bounds. This as an avenue for future work, which is likely associated with an analysis of BOND's upper bound as an optimizer, given its proximity to Adam in Table~\ref{table:bondvsadam_sign}.

\section{Conclusion}
We introduce BOND, a zeroth-order method that enables end-to-end training of hybrid architectures combining neural networks with black-box functions. BOND leverages local computational graphs and employs adaptively bounded perturbations to achieve convergence comparable to automatic differentiation. Unlike existing methods which scale with parameter space, BOND scales only with black-box input dimensionality. Using the California House Price, FashionMNIST, CIFAR10 and CIFAR100 datasets, we demonstrate that embedding black-box functions in neural networks \textit{can} improve performance without increasing trainable parameters. Although limitations remain, this work establishes a foundation for integrating non-autodifferentiable components into modern neural networks and raises important questions about the role of untrained transformations in deep learning.

\newpage
\newpage
\bibliography{references}

\onecolumn
\newpage
\appendix

\section{Additional Related Work}\label{appendix:related_works}
\subsection{Zeroth-Order Methods}
We note that SPSA does permit some control of the estimation-accuracy trade-off, using the hyperparameter $n_{\textit{pert}}$ to allow for additional forward passes, each with a new perturbation vector. Averaging the finite-difference computations across these estimates creates a more reliable gradient approximator, which improves performance, however partially negates the SPSA computational advantage.

\citet{pgt} introduce a SPSA variant which seeks to increase the number of search directions by modifying perturbation magnitudes, however we find this method comparable to conventional SPSA. In Equation \ref{eq:central_spsa}, we illustrate a standard central-difference implementation of SPSA, which is modified from \textit{Eq. 3} of \citet{SPSA2001SpallOverview}.
\begin{equation}\label{eq:central_spsa}
    {\hat{\nabla}_{t,i} f(\theta_{t})}_{SPSA} = \frac{1}{n_{\textit{pert}}} \sum_{k=1}^{n_{\textit{pert}}} \frac{f(\theta_t + \mu_t \Delta_{t,k}) - f(\theta_t - \mu_t \Delta_{t,k})}{2\mu_t \Delta_{t,k,i}}
\end{equation}
where $\Delta_{t,k}$ is a random perturbation vector sampled from a distribution $\mathcal{P}$ at iteration $t$ and perturbation $k$, and $\mu_t$ is a global smoothing coefficient. The notation $i$ is used to denote the $i$'th component of $\hat{\nabla}_{t} f(\theta_{t})$ ($i=$1,2, ..., $d_{\theta}$). In Equation~\ref{eq:central_fdsa}, we show a central-difference FDSA formulation.
\begin{equation}\label{eq:central_fdsa}
    {\hat{\nabla}_{t,i} f(\theta_{t})}_{FDSA} = \frac{f(\theta_t + \mu_t \Delta_t e_i) - f(\theta_t - \mu_t \Delta_t e_i)}{2\mu_t \Delta_{t,i}}
\end{equation}
Here, $e_i$ is a unit-vector with a one in the $i$'th component and zero elsewhere. We note the forward-difference implementation of Equation \ref{eq:central_fdsa} would be defined with the numerator $f(\theta_t + \mu_t \Delta_t e_i) - f(\theta_t)$, however consider that the central-difference method is preferable as it incurs less variance, noted in \textit{Eq. 3} of \citet{scaling_rnns}, and in \citet{nesterov_gradfree_convergences}. 

\citet{zo_adam} and \citet{duchi_convergence} show that the convergence rate of SPSA is approximately $\mathcal{O}(\sqrt{d_{\theta}})$ worse than FO methods. \citet{zo_adam}, in \textit{Proposition 1}, also uses the special case $\beta_1,\beta_2\rightarrow0$ to demonstrate non-convergence issues with euclidean projection in parameter updates, highlighting the case for ZO optimization with Adam. We note that \citet{duchi_convergence}, in \textit{Corollary 2}, identifies a trade-off between the estimation complexity $\mathcal{O}(n_{pert})$ and the relative rate of convergence, given by a factor $\mathcal{O}(\sqrt{1+\frac{d_{\theta}}{n_{pert}}})$. For the limit $n_{pert}=d_{\theta}$, this suggests that SPSA achieves a similar convergence rate to FO methods at the cost of $d_{\theta}$ function evaluations per iteration. \citet{fdsa_convergence} similarly demonstrates that common random number sequences can be used with FDSA to approach FO convergence rates, however, this is grounded on the assumption that $\theta$ is a scalar, and does not comment on estimation complexity. These works also depend on the assumption of smoothness as $\mu_t \rightarrow 0$, which \textit{can} be unrealistic in practice with physical reservoir computing devices, a notion which is further discussed in Appendix \ref{appendix:perturbation_magnitudes}.

Additionally, we adopt a uniform distribution $\mathcal{P}$ for sampling perturbation size, following \citet{signSGD_liu,zo_adam}, preferring its boundedness properties for convergence and formulation over the Normal and Rademacher distributions advocated by \citet{scaling_rnns}.

We also note a parallel literature series on perturbative gradient estimation \cite{node_perturbation_source_dembo,node_perturbation_source_gert}, developed independently of SPSA \cite{Spall_og,robbinsmonro1951stochastic} despite significant conceptual overlap. This work serves as an early foundation for Node Perturbation \citep{node_perturbation,node_perturbation_stability} — effectively an SPSA variant related to BOND in its perturbation of nodes rather than parameters. A key technique from this literature is the normalization of gradient estimates to mitigate variance.

\subsection{Reservoir Computing} 
Conventional examples of RC include Echo State Networks (ESN) \cite{jaeger_tute} and liquid state machines \cite{wolfganggang}, which employ recurrent connections, allowing for memory retention. This property is especially useful for problems with sequential dynamics such as time-series prediction \cite{LUKOSEVICIUS2009127,jaeger2} and speech recognition \cite{10.1016/j.neunet.2007.04.016, 1716215}. However, the computation defined by physical RC devices can be fixed or time-dependent in nature, suggesting that the utility of a physical device in efficiently offloading digitized computation holds promise for both applications with and without sequential dynamics. Existing explorations of physical devices in RC include examples like a literal bucket of water in \cite{bucketwater,en16145366}, though they more commonly involve electronic, mechanical, biological or chemical structures \cite{vrugt2024introductionreservoircomputing,TANAKA2019}. More intricate designs include the development of neuromorphic spiking networks in \cite{neuromorphic, neuromorphic_structured}, and quantum reservoirs \cite{fujii2020quantumreservoircomputingreservoir}, supported by recent applications of quantum-based circuits as feature transforms \cite{hsbc,quantum_spin_timeseries}. Although each method independently merits further investigation, BOND's unique access to accurate derivative information enables novel research directions, including the integration of such devices into trainable neural network architectures.

\subsection{Embedding black-box functions in trainable architectures}
We find limited existing research on embedding black-box functions in trainable architectures. \citet{multiplex} explores ZO methods with MNIST and CIFAR10, achieving results which are promising, but not competitive with AD. Using transformers, \citet{pgt} also demonstrates partial convergence with a physical RC device, however, their results are constrained by SPSA-type gradient error and the complexity of transformer integration. While unrelated to gradient estimation, Random Projection (RP) layers \cite{rand_projections} parallel our use of a black-box function for efficient linear separability, with relevant observations on non-linearity, rank improvement, and correlated performance gains. We also note that existing literature broadly supports embedding black-box functions within this framework. ZO optimization has proven effective under masked or complex dynamics \cite{liu2020primer,yan2024emerging}, while the success of scale-invariant \citep{adam} and sign-based \citep{signSGD_liu,lion} optimizers suggests gradient magnitude errors are not fundamental to convergence \citep{zo_adam}, motivating our focus on gradient sign accuracy.

\section{Support for BOND formulations}\label{appendix:proofs}

\subsection{Assumptions}

\begin{assumption}[Lipschitz Continuity]
\label{assump:diff_3}
The function $f$ has a bounded $L$-Lipschitz continuous third derivative i.e. for all $x, y \in \mathcal{X}$ and $L_2,L_1,L_0 \in(0,\infty)$:
\[
\|f(x) - f(y)\| \leq L_0\|x - y\|
\]
\[
\|f'(x) - f'(y)\| \leq L_1\|x - y\|
\]
\[
\|f''(x) - f''(y)\| \leq L_2\|x - y\|
\]
\end{assumption}

\begin{assumption}[Uniform Noise Distribution]
\label{assumption:uniform_noise}
The noise $\epsilon$ in the function $f^{\star}(x) = f(x) + \epsilon$ is drawn from an unknown and unbiased uniform distribution, i.e. $\epsilon \overset{\text{iid}}{\sim} U(-b,b)$.

\end{assumption}
We contend that uniform baseline noise is reasonable as errors determined by floating point operations are bounded~\citep{noise_boundedness}, and a uniform distribution is the limit for the worst case bounded normal distribution. 

\subsection{Gradient approximation}
Consider the central-difference approximation $(g_t)_i$ of the gradient $\nabla_i f_{t}(x_t)$ at dimension $i$ for iteration $t$, with access to the function $f_{t} : \mathbb{R}^{d} \rightarrow \mathbb{R}$.

\begin{equation}
    (g_t)_i = \frac{f_{t}(x_t + \Delta_{x}^t e_i) - f_{t}(x_t - \Delta_{x}^t e_i)}{2(\Delta_{x}^t)_i}
\end{equation}

Using a 2nd order Taylor approximation of $f_t(x_t \pm \Delta_{x}^t e_i)$ centered at $x_t$, we have:
\[
    f_{t}(x_t + \Delta_{x}^t e_i) = f_{t}(x_t) + \Delta_{x}^t e_i \nabla f_{t}(x_t) + \frac{1}{2} (\Delta_{x}^t e_i)^2 \nabla^2 f_{t}(x_t) + R_2^{+}
\]
\[
    f_{t}(x_t - \Delta_{x}^t e_i) = f_{t}(x_t) - \Delta_{x}^t e_i \nabla f_{t}(x_t) + \frac{1}{2} (\Delta_{x}^t e_i)^2 \nabla^2 f_{t}(x_t) + R_{2}^{-}
\]
\begin{align} \label{grad_with_remainder}
    \therefore (g_t)_i = \nabla_i f_{t}(x_t) + \frac{(R_2^{+} - R_2^{-})}{2(\Delta_{x}^t)_i}
\end{align}
Where $R_2$ is the Lagrange remainder term for the 2nd order Taylor expansion:
\[
    R_2^{\pm} = \frac{1}{6} (\nabla_{i}^3 f_t(x_t + c \Delta_{x}^t e_i))((\Delta_{x}^t)_i)^3
\]
Where $c \in [-1, 1]$ is unknown, and $\nabla_{i}^3 f_t$ is the third derivative of $f_t$ in the $i$-th dimension. Typical analyses set error bounds based on the supremum $(M_t)_i$ of the third derivative across the interval $[x \pm (\Delta_{x}^t)_i]$. Using Assumption~\ref{assump:diff_3}, we define $L_2$ as the Lipschitz constant bounding the supremum of the third derivative over any interval:
\begin{align}
    \lvert R_2^{\pm} \rvert \leq \frac{L_2}{6} \lvert ((\Delta_{x}^t)_i)^3 \rvert
\end{align}

This enables us to bound the difference with the triangle inequality to get: $\lvert R_2^+-R_2^-\rvert \leq \frac{L_2}{3} \lvert((\Delta_{x}^t)_i)^3\rvert$. Substituting into and rearranging \ref{grad_with_remainder}

\begin{align}
    \lvert (g_t)_i - \nabla_i f_t(x_t)\rvert 
    & = \frac{1}{2}\Big\lvert \frac{R_2^+-R_2^-}{((\Delta_{x}^t)_i)} \Big \rvert \notag \\
    \therefore \lvert (g_t)_i - \nabla_i f_t(x_t)\rvert 
    & \leq \frac{L_2}{6} \lvert ((\Delta_{x}^t)_i)^2 \rvert \label{eq:smooth_error}
\end{align}

\subsection{Noisy function}
In Equation~\ref{eq:smooth_error}, it is intuitive that the assumption $(\Delta_{x}^t)_i \rightarrow 0$ (i.e. $\mu_t \rightarrow 0$) creates the most accurate gradient approximation. However, in practice, underlying noise characteristics may prohibit the validity of this assumption. For example, if $x=1$ and $\Delta_{x}^t=1e-10$, computing with the numerical estimate fp32 will likely not lead to an accurate approximation. We introduce $f_{t}^{\star}(x)$ as the noisy analog of $f_{t}$, defined as:
\[
    f_{t}^{\star}(x) = f_{t}(x) + \epsilon
\]

Using the same 2nd order Taylor series approximation on $f_{t}^{\star}(x)$, we have:

\[
    (g^{\star}_t)_i = \frac{f_{t}^{\star}(x_t + \Delta_{x}^t e_i) - f_{t}^{\star}(x_t - \Delta_{x}^t e_i)}{2(\Delta_{x}^t)_i}
\]

Similarly, following the working for Equation~\ref{eq:smooth_error}, using the fact that $\lvert\epsilon_{+} - \epsilon_{-}\rvert \leq 2b$ from Assumption~\ref{assumption:uniform_noise}. 

\[
    \because f_{t}^{\star}(x_t \pm \Delta_{x}^t e_i) = f_{t}(x_t \pm \Delta_{x}^t e_i) + \epsilon_{\pm}
\]
\begin{align}
    \therefore \lvert (g_t^{\star})_i - \nabla_i f_t(x_t)\rvert 
    & = \frac{1}{2}\Big\lvert \frac{R_2^+-R_2^-}{((\Delta_{x}^t)_i)} 
    + \frac{\epsilon_+ - \epsilon_-}{((\Delta_{x}^t)_i)} 
    \Big \rvert \notag
\end{align}
\begin{align}
    \therefore \lvert (g_t^\star)_i - \nabla_i f_{t}(x_t)\rvert 
    \leq \frac{L_2}{6} \lvert ((\Delta_{x}^t)_i)^2 \rvert 
    + b \big\lvert \frac{1}{(\Delta_{x}^t)_i} \big\rvert \label{eq:grad_error}
\end{align}

We define the error norm relative to the gradient norm:
\begin{align}
    \frac{\lvert (g_t^\star)_i - \nabla_i f_{t}(x_t)\rvert}{\lvert \nabla_i f(x_t) \rvert} 
    \leq \frac{1}{\lvert \nabla_i f_{t}(x_t) \rvert} \left[ \frac{L_2}{6} \lvert ((\Delta_{x}^t)_i)^2 \rvert 
    + b \big\lvert \frac{1}{(\Delta_{x}^t)_i} \big\rvert \right] \label{eq:relative_gradient_error}
\end{align}

In \ref{eq:relative_gradient_error}, the individual quadratic/sublinear component functions of the upper bound respectively monotonically increase/decrease in $(\Delta_{x}^t)_i$. These dominate for different domains of $(\Delta_{x}^t)_i$ and intersect at
\[
    \frac{L_2}{6} \lvert ((\Delta_{x}^t)_i)^2 \rvert 
    = b \big\lvert \frac{1}{(\Delta_{x}^t)_i} \big\rvert
\]
\[
    \therefore \lvert (\Delta_{x}^t)_i \rvert = \sqrt[3]{\frac{6b}{L_2}}
\]

We use this to define two regions of interest for $(\Delta_{x}^t)_i$ which we analyze further in the below sections.

\subsection{Lower and Upper Bounds for \texorpdfstring{$(\Delta_{x}^t)_i$}{Delta}}

This section will explore the various regions in which $(\Delta_{x}^t)_i$ can lie based on the amount of noise $b$ in the function $f^{\star}$.

The $b$ term dominates for  $\lvert (\Delta_{x}^t)_i \rvert < \sqrt[3]{\frac{6b}{L_2}}$, and by substituting this into \ref{eq:relative_gradient_error} gives:
\begin{align}
    \frac{\lvert (g_t^\star)_i - \nabla_i f_{t}(x_t)\rvert}{\lvert \nabla_i f_{t}(x_t) \rvert} \leq \frac{1}{\lvert \nabla_i f_{t}(x_t) \rvert}\left[ \frac{L_2}{6} \Big(\sqrt[3]{\frac{6b}{L_2}} \Big)^2 + \frac{b}{|(\Delta_{x}^t)_i|} \right] \notag
\end{align}

Remark: the above is the highest upper bound on the relative error in the region where $\lvert(\Delta_{x}^t)_i\rvert < \sqrt[3]{\frac{6b}{L_2}}$.

In this region, if the upper bound on the relative error is no greater than 1, then the gradient sign is correct:

\begin{align}
    \frac{1}{\lvert \nabla_i f_{t}(x_t) \rvert}\left[ \frac{L_2}{6} \Big(\sqrt[3]{\frac{6b}{L_2}} \Big)^2 + \frac{b}{|(\Delta_{x}^t)_i|} \right] 
    \leq 1 \implies \text{sgn}(g_t^\star)_i = \text{sgn}(\nabla_i f_{t}(x_t))
    \notag
\end{align}

We use this fact to derive a lower bound for $(\Delta_{x}^t)_i$ in this region. 

Let $C_t = \frac{L_2}{6} \Big(\sqrt[3]{\frac{6b}{L_2}} \Big)^2$ for brevity. On rearranging the above inequality and maintaining the gradient sign correctness condition gives:

\begin{align}
    C_t + \frac{b}{\lvert(\Delta_{x}^t)_i\rvert} \leq \lvert \nabla_i f_{t}(x_t) \rvert \notag\\
    \frac{b}{\lvert \nabla_i f_{t}(x_t) \rvert - C_t} \leq  \lvert(\Delta_{x}^t)_i\rvert \label{eq:lower_bound}
\end{align}

Heuristically, we have the following interpretations of the perturbation lower bound.

As $b \rightarrow 0$:
\[
    C_t \rightarrow 0, \text{ and } 0 \leq \lvert(\Delta_{x}^t)_i \rvert
\]
When $b > \frac{L_2}{6}\Big(\sqrt{\frac{6\lvert \nabla_i f_{t}(x_t) \rvert}{L_2}}\Big)^3$
\[
    \text{No valid solution}
\]
since the denominator becomes negative, and $\lvert(\Delta_{x}^t)_i \rvert \geq 0$ by definition.

We find these outcomes to be intuitive, as no noise implies that the assumption $\mu_t~\rightarrow~0$ holds, and would naturally lead to the most precise estimation of the instantaneous gradient. Additionally, for sufficiently large noise, there may not be any perturbation range which satisfies the condition of relative error being less than 1.
\\

Similarly, for the region $\lvert(\Delta_{x}^t)_i\rvert > \sqrt[3]{\frac{6b}{L_2}}$, and using the substitution $D_t = \frac{b}{\sqrt[3]{\frac{6b}{L_2}}}$:
\begin{align}
    \frac{\lvert (g_t^\star)_i - \nabla_i f_{t}(x_t)\rvert}{\lvert \nabla_i f_{t}(x_t) \rvert}  &\leq \frac{1}{\lvert \nabla_i f_{t}(x_t) \rvert}\Big[ \frac{L_2}{6} ((\Delta_{x}^t)_i)^2 + D_t \Big] \notag \\
    \Big[ \frac{L_2}{6} ((\Delta_{x}^t)_i)^2  + D_t \Big] &\leq \lvert \nabla_i f_{t}(x_t) \rvert \notag \\
    \lvert (\Delta_{x}^t)_i \rvert &\leq \sqrt{\frac{6}{L_2}\big(\lvert \nabla_i f_{t}(x_t) \rvert - D_t \big)} \label{eq:upper_bound}
\end{align}

Where we examine the case as $b \rightarrow 0$:
\[
    D_t \rightarrow 0, \text{ and } \lvert (\Delta_{x}^t)_i \rvert \leq \sqrt{\frac{6 \lvert \nabla_i f_{t}(x_t) \rvert}{L_2}}
\]
We find this case to be reasonable and compatible with the previous findings for the lower bound.
\\
\\
However, we must make the practical considerations of what information we have access to when determining the perturbation magnitudes (i.e. $(\Delta_{x}^t)_i$ must be determined prior to $\nabla_i f_t(x_t)$ being known $-$ and therefore cannot be formulated from it). At iteration $t$, given $x_t$, we use a uniform distribution $\epsilon_{\nabla}^t \overset{\text{iid}}{\sim} U(a^{t-1}_-,a^{t-1}_+)$ to quantify the change in the true gradient based on parameter updates and data across iterations. Unlike $\epsilon_{\pm}$, $\epsilon_{\nabla}^t$ will reduce as the functions converges, making it iteration dependent. We also don't have access to the true Lipschitz constant bounding the third derivative $\nabla_{i}^3 f_t(.)$. Hence, let $L^{t}_2$ be the estimate of $L_2$ at iteration $t > 0$ which uses information from the rolling estimates across previous iterations $\hat{v}_{t-1}$ (see Algorithm \ref{pseudo_code_rollingmoments}).

We have
\[
    \lvert \nabla_i f_{t}(x_t) - \nabla_i f_{t-1}(x_{t-1}) \rvert = \lvert \epsilon_{\nabla}^t \rvert
\]
Using the triangle inequality, to bound the error terms.

\[
    \lvert \nabla_i f_{t}(x_t) - (g_{t-1}^\star)_i \rvert 
    \leq \lvert \nabla_i f_{t}(x_t) - \nabla_i f_{t-1}(x_{t-1}) + \nabla_i f_{t-1}(x_{t-1}) - (g_{t-1}^\star)_i \rvert 
\]

\[
    \lvert \nabla_i f_{t}(x_t) - (g_{t-1}^\star)_i \rvert 
    \leq \lvert \nabla_i f_{t}(x_t) - \nabla_i f_{t-1}(x_{t-1}) \rvert 
    + \lvert \nabla_i f_{t-1}(x_{t-1}) - (g_{t-1}^\star)_i \rvert 
\]
Using Equation \ref{eq:grad_error}.
\[
    \lvert \nabla_i f_{t}(x_t) - (g_{t-1}^\star)_i \rvert 
    \leq \lvert \epsilon_{\nabla}^t \rvert
    + \frac{L^{t-1}_2}{6} ((\Delta_{x}^{t-1})_i)^2 
    + \frac{b}{\lvert(\Delta_{x}^{t-1})_i\rvert}
\]

\[
    \lvert \nabla_i f(x_{t}) \rvert
    \in \left[
        \lvert (g_{t-1}^\star)_i \rvert
        \pm \left\lvert 
            \frac{L^{t-1}_2}{6} ((\Delta_{x}^{t-1})_i)^2 
            + \frac{b}{\lvert(\Delta_{x}^{t-1})_i\rvert}
            + a^{t-1} 
        \right\rvert
    \right]
\]
Given that Equation \ref{eq:lower_bound} sets the lower bound, and the term $\lvert \nabla_i f_{t}(x_t)\rvert$ is in the denominator, we must consider the lower bound's infimum. This occurs with the smallest possible value of $\lvert \nabla_i f_{t}(x_t)\rvert$ in the interval constructed above. This leads to the formulation:

\begin{align}
    \frac{b}{\lvert (g_{t-1}^\star)_i \rvert
    - \lvert \frac{L^{t-1}_2}{6} ((\Delta_{x}^{t-1})_i)^2 
    + \frac{b}{\lvert(\Delta_{x}^{t-1})_i\rvert}
    + a^{t-1} \rvert 
    - \frac{L^{t-1}_2}{6} \Big(\sqrt[3]{\frac{6b}{L^{t-1}_2}} \Big)^2} \leq  \lvert(\Delta_{x}^t)_i\rvert \label{eq:lower_bound_full}
\end{align}
With the same rationale, we also substitute the infimum into Equation \ref{eq:upper_bound}, setting the smallest possible upper bound.
\begin{align}
    \lvert(\Delta_{x}^t)_i \rvert&\leq \sqrt{\frac{6}{L^{t-1}_2}
    \left(\lvert (g_{t-1}^\star)_i \rvert
    - \lvert \frac{L^{t-1}_2}{6} ((\Delta_{x}^{t-1})_i)^2 
    + \frac{b}{\lvert(\Delta_{x}^{t-1})_i\rvert}
    + a^{t-1} \rvert  - \frac{b}{\sqrt[3]{\frac{6b}{L^{t-1}_2}}} \right) }
    \label{eq:upper_bound_full}
\end{align}

However, equations \ref{eq:lower_bound_full} and \ref{eq:upper_bound_full} are not easily implemented, involving the terms $a^{t-1}$ and $L_2^{t-1}$ which require additional estimations. In this context, we simplify the above via a series of heuristic and empirically motivated substitutions. Despite providing rationale for all modifications, we acknowledge that an improved implementation of equation \ref{eq:lower_bound_full} and \ref{eq:upper_bound_full} is an important focus for future work. From Figure~\ref{fig:stacked}, we find that removing the extra negative term $- (\lvert \frac{L^{t-1}_2}{6} ((\Delta_{x}^{t-1})_i)^2 + \frac{b}{\lvert(\Delta_{x}^{t-1})_i\rvert} + a^{t-1} \rvert  + \frac{b}{\sqrt[3]{\frac{6b}{L^{t-1}_2}}})$ has a small but positive effect on the upper bound formulation, generally leading to slightly larger bounds. This is likely influenced by the imprecise estimations of $a^{t-1}$ and $L_2^{t-1}$.  We also find the constant $6$ has a negligible impact in performance, and can be dropped. This leads to the revised bounds.
\begin{align}
    \overline{(\Delta_{x}^t)_i} 
    &\leq \sqrt{\frac{\lvert (g_{t-1}^\star)_i \rvert }{L_2^{t-1}}} \label{eq:upper_bound_revised}\\
    \underline{(\Delta_{x}^t)_i} 
    &\geq \frac{b}{\lvert (g_{t-1}^\star)_i \rvert} \label{eq:lower_bound_revised}
\end{align}

We find Equation~\ref{eq:lower_bound_revised} to be suitable in implementation, however note that we include a modification, whereby we take the minimum over $\lvert (g_{t-1}^\star)_i \rvert$ which creates a slightly larger lower bound, providing more consistent performance. 

Equation~\ref{eq:upper_bound_revised} requires additional work. As noted above, the estimation of $L_2^{t-1}$ is non-trivial, as we do not have access to third derivative information. Therefore, we propose 2 key modifications to this formula that are heuristically motivated rather than theoretically grounded. We introduce a scaling factor $\alpha_{t}=\max{\lvert (g_{t-1}^\star)_i \rvert}$, which bounds the perturbations similarly to a learning rate in a parameter update, and bolsters the coupling effect that leads to equilibrium, i.e.
$(\Delta_{x}^{t-1})_i \uparrow \implies \max\lvert (g_{t-1}^\star)_i \rvert \downarrow \implies (\Delta_{x}^{t})_i \downarrow \implies \max\lvert (g_{t}^\star)_i \rvert \uparrow$. 
Secondly, as $L_2^{t-1}$ requires third-order information, we replace it with an estimate based on second-order information. Since Equation~\ref{eq:upper_bound_revised} is decreasing in $L_2^{t-1}$, substituting any $L \geq L_2^{t-1}$ preserves the bound. We therefore use the gradient Lipschitz constant $L_1^{t-1}$, assuming $L_1^{t-1} \geq L_2^{t-1}$, which holds when the objective is locally well-approximated by a quadratic, but is not guaranteed in general. Following the interpretation of Adam's second moment as a diagonal Hessian approximation, we estimate $L_1^{t-1} \approx \max (\hat v^{\mathcal R}_{t-1})$, and replace $g^\star_{t-1}$ with its rolling moment estimation  $\hat m^{\mathcal R}_{t-1}$.
\begin{align}
    \overline{(\Delta_{x}^t)_i} 
    &\leq \max{\lvert (\hat{m}^{\mathcal{R}}_{t-1}) \rvert} \cdot \sqrt{\frac{\lvert (\hat{m}^{\mathcal{R}}_{t-1})_i \rvert } {\max{(\hat{v}^{\mathcal{R}}_{t-1})}}} \label{eq:upper_bound_revised_revised}
\end{align}
Replacing $\max(\hat v^{\mathcal R}_{t-1})$ with the coordinate-wise $(\hat v^{\mathcal R}_{t-1})_i$ forgoes the global scaling in favour of a per-coordinate curvature estimate, which yielded slightly improved performance in initial testing. We also introduce $\epsilon$ for numerical stability.
\begin{align}
    \overline{(\Delta_{x}^t)_i} 
    &\leq \max{\lvert (\hat{m}^{\mathcal{R}}_{t-1}) \rvert} \cdot \sqrt{\frac{\lvert (\hat{m}^{\mathcal{R}}_{t-1})_i \rvert } {(\hat{v}^{\mathcal{R}}_{t-1})_i + \epsilon } } \label{eq:upper_bound_}
\end{align}

Whilst these bounds do achieve reasonable performance, we recognize that additional work is required to fully establish a more rigorous set of implementable bounds from Equation~\ref{eq:upper_bound_full}. This would enable the generalization of BOND to functions where the current formulations may fail. Specifically, the substitutions used to obtain Equation~\ref{eq:upper_bound_} do not preserve the dimensional consistency of Equation~\ref{eq:upper_bound_full}, such that the implemented bound may be sensitive to the scaling of the loss and black-box inputs. \\

We also acknowledge that another general approach to determining perturbation magnitude would be to take the derivative of the absolute error bound with respective to perturbation magnitude. This would require that ${(\Delta_{x}^{t-1})_i}$ be a positive constant, instead of being randomly sampled between the bounds based on noise. Setting the derivative to zero would give the optimal perturbation magnitude which minimizes the error bound. This gives the below.
\begin{align}
    \lvert (g_t^\star)_i - \nabla_i f_{t}(x_t)\rvert
    &\leq \frac{L^{t}_2}{6} {(\Delta_{x}^t)_i}^2
    + \frac{b}{{(\Delta_{x}^t)_i}} \notag
\end{align}
\begin{align}
    \frac{d}{d{(\Delta_{x}^t)_i}} \Big(\frac{L_2^{t}}{6} {(\Delta_{x}^t)_i}^2
    + \frac{b}{{(\Delta_{x}^t)_i}} \Big) &= 0 \notag \implies
    (\Delta_{x}^t)_i = \sqrt[3]{\frac{3b}{L_2^{t}}}
\end{align}
\paragraph{Remark:} we explored both approaches with random sampling and fixed perturbations. Likely due to the erroneous $L_2^t$ approximations, we found that random sampling provided robusticity and enabled more consistent performance. Improving this approximation may yield a more computationally efficient BOND application.

\newpage
\section{Algorithms}\label{appendix:algorithms}
The BOND algorithm is detailed below, along with supporting algorithms for computing adaptive perturbation bounds and gradient moment estimates.

\subsection{BOND Algorithm}
The set of input-output training samples are denoted $X$ and $Y$. We use three sequential networks, defined as $f_A(\cdot,\,\theta_{A})$, $\mathcal{R}(\cdot)\equiv\mathcal{R}(\cdot,\,\theta_{\mathcal{R}})$, and $f_B(\cdot,\,\theta_{B})$, with the assumption that $f_A(\cdot,\theta_{A})$ and $f_B(\cdot,\theta_{B})$ have separate but well-defined computational graphs. $\mathcal{R}(\cdot)$ is a smooth black-box function, however we do not require that its computational graph is accessible. Naturally, in our black-box simulations (ESN's and FFNN's) we have access to the $\mathcal{R}(\cdot)$ graph, enabling comparisons between numerical estimations and calculated gradients. The loss function $\mathcal{L}$ (Huber Loss \& Cross Entropy) is standard and the choice of optimizer is flexible, i.e. we track the rolling moments of reservoir inputs independently of optimizer implementation. We provide the pseudocode in Algorithm~\ref{pseudo_code_bond}. The terms $\hat{m}^{\mathcal{R}}_{t-1},\hat{v}^{\mathcal{R}}_{t-1}$ are standard rolling estimates, defined by Algorithm \ref{pseudo_code_rollingmoments} in Appendix \ref{appendix_rollingmoments}. The function $\texttt{setPerturbations}$ determines the perturbations using Equation \ref{eq:lower_bound_used} and \ref{eq:upper_bound_used}, and is shown in Algorithm \ref{pseudo_code_setnoise} Appendix~\ref{appendix_set_perturbations}.

\begin{algorithm2e}[H]
    \caption{BOND [central-difference]}
    \label{pseudo_code_bond}
    \SetAlgoVlined
    $Y_A \gets f_{A}(X,\theta_{A})$\;

    $Y_{\mathcal{R}} \gets \mathcal{R}(Y_A)$\;

    $Y_B \gets f_{B}(Y_{\mathcal{R}} ,\theta_{B})$\;

    $L\gets \mathcal{L}(Y,Y_B)$\;

    $\frac{\partial L}{\partial \theta_B} \gets L.\texttt{Backward()}$\;

    $\Delta_x^t \gets \texttt{setPerturbations(}\hat{m}^{\mathcal{R}}_{t-1},\hat{v}^{\mathcal{R}}_{t-1}, b \texttt{)}$\;

    \For{$i \in \{0,1,...,d_{\mathcal{R}}-1\}$}{
        ${Y_{\mathcal{R}}^-}_{i} \gets \mathcal{R}(Y_{A}- \Delta_{x}^t e_i)$\;

        ${Y_{\mathcal{R}}^+}_{i} \gets \mathcal{R}(Y_{A}+ \Delta_{x}^t e_i)$\;
    }

    $\hat{\frac{\partial Y_{\mathcal{R}}}{\partial Y_{A}}} = \frac{Y_{\mathcal{R}}^+ -Y_{\mathcal{R}}^-}{2\Delta_x^t}$\;

    $\hat{\frac{\partial L}{\partial \theta_{A}}} \gets Y_{A}.\texttt{Backward(gradient = } \frac{\partial L}{\partial Y_{\mathcal{R}}} \cdot \hat{\frac{\partial Y_{\mathcal{R}}}{\partial Y_{A}}} \texttt{)}$\;
\end{algorithm2e}


\subsection{Set Perturbations function}\label{appendix_set_perturbations}
\begin{algorithm2e}[H]
    \caption{$\texttt{setPerturbations}$}
    \label{pseudo_code_setnoise}
    \SetAlgoVlined
    \If{$t$ is 0}{
        $z \sim \mathcal{N}(0,\sqrt{b})$\;

        $\Delta_x^t \gets \text{sign}(z) \cdot \max\{|z|, b\}$\;
    }
    \Else{
        $\alpha_t \gets \max\big\{|\hat{m}^{\mathcal{R}}_{t-1}|\big\} $\;

        $\overline{\Delta_x^t} = \alpha_{t} \cdot \sqrt{|\hat{m}^{\mathcal{R}}_{t-1}| / (\hat{v}^{\mathcal{R}}_{t-1}+\epsilon)}$\;

        $\underline{\Delta_x^t} \gets b / (\min \big\{|\hat{m}^{\mathcal{R}}_{t-1}|\big\} + \epsilon)$\;

        $\lvert\Delta_x^t\rvert \sim \mathcal{U} [\underline{\Delta_x^t},\overline{\Delta_x^t}]$\;

        $s \sim \{+1,-1\}^{\lvert\Delta_x^t\rvert}$\;

        $\Delta_x^t \gets s \cdot \lvert\Delta_x^t\rvert$\;
    }
\end{algorithm2e}
Note that in Algorithm~\ref{pseudo_code_setnoise}, the absolute value of $\hat{m}^{\mathcal{R}}_{t-1}$ ensures non-negativity under the root. We use $\epsilon=1e-8$ conventionally ensure numerical stability. Additionally, we randomize the signs of the pertubations, such that there is no form of consistent positive/negative bias when applying the central-difference peturbations.

\subsection{Rolling gradient estimates}\label{appendix_rollingmoments}
The rolling moment estimation used for BOND is equivalent to that in Adam \citep{adam}, just applied to reservoir inputs rather than network parameters. We find that using these rolling moments results in better perturbation magnitudes compared to using the direct gradient estimates (i.e. $\beta_1,\beta_2 \rightarrow 0$). To enable compatibility with any optimizer, we implement this algorithm independently for each iteration. We use the standard $\beta_1=0.9$ and $\beta_2=0.999$ in our implementations, where no grid search was performed outside these values.

\begin{algorithm2e}[H]
    \caption{Rolling moment estimates}
    \label{pseudo_code_rollingmoments}
    \SetAlgoVlined
    $m^{\mathcal{R}}_t \gets \beta_1 m^{\mathcal{R}}_{t-1} + (1 - \beta_1) g_t$\;

    $v^{\mathcal{R}}_t \gets \beta_2 v^{\mathcal{R}}_{t-1} + (1 - \beta_2) g_t^2$\;

    $\hat{m}^{\mathcal{R}}_t \gets \frac{m^{\mathcal{R}}_t}{1 - \beta_1^t}$\;

    $\hat{v}^{\mathcal{R}}_t \gets \frac{v^{\mathcal{R}}_t}{1 - \beta_2^t}$\;
\end{algorithm2e}

\subsection{BONDS}\label{appendix_bonds}
BONDS is a simple addendum to the original BOND Algorithm, see Algorithm~\ref{pseudo_code_bond}, replacing lines $7-10$ with Algorithm~\ref{pseudo_code_bonds}. This is effectively just swapping the for loop with the simultaneous implementation, and adding optionality for row normalization from Equation \ref{eq:row_normalization}. We make the note that, for BONDS, the row normalization default is $normalize=False$, but we frequently include it in initial explorations.

\begin{algorithm2e}[H]
    \caption{BONDS Addendum}
    \label{pseudo_code_bonds}
    \SetAlgoVlined
    ${Y_{\mathcal{R}}^-},{Y_{\mathcal{R}}^+} \gets \mathcal{R}(Y_{A}- \Delta_{x}^t),\mathcal{R}(Y_{A}+\Delta_{x}^t)$\;

    $\hat{g} = \frac{Y_{\mathcal{R}}^+ -Y_{\mathcal{R}}^-}{2\Delta_x^t}$\;

    \If{normalize}{
        $\hat{g}_{k} =  \frac{\hat{g}_{k}}{\|\hat{g}_{k,:}\|}$\;
    }
\end{algorithm2e}

Algorithm~\ref{pseudo_code_bonds} includes the row normalization which is defined in Equation~\ref{eq:row_normalization}
\begin{equation}\label{eq:row_normalization}
\hat{g}_{k} = \frac{\hat{g}_{k}}{\|\hat{g}_{k,:}\|}, \ \forall k \in \mathcal{B} 
\end{equation}
Where $k$ is a sample index from the batch $\mathcal{B}$. For recursive and simultaneous perturbations, gradient estimates exhibit higher row norm variability, motivating a simple row normalization Equation \ref{eq:row_normalization}. This is further grounded by \citet{equilibration} \textit{Proposition 4}, which shows row normalization reduces the upper bound on the Hessian condition number — consistent with our observation that uniform row norms reduce its efficacy in NFN-BOND.

\newpage
\section{Additional Results}\label{appendix:additional_results}
\subsection{Califronia House Price Results}\label{appendix:chp_results}
This table contains training information for performance on the California House Price Dataset for NEN and NFN achitectures. Corresponding to the plots in Figure~\ref{fig:calhouse_loss}.
\begin{table}[ht]
    \caption{Training Information - California House Price}
    \label{table:calhouse}
    \centering
    \begin{tabular}{lrr}
        \toprule
        Model & Loop Time (ms)& Correct Sign (\%) \\
        \midrule
        NEN - AD & $ 16.57 \pm \scriptstyle 1.75$ & $-$ \\
        NEN - BOND & $ 91.82 \pm \scriptstyle 12.41$ & $79.86 \pm \scriptstyle 20.63$ \\
        NEN - BONDS & $ 90.64 \pm \scriptstyle 10.11$ & $60.81 \pm \scriptstyle 14.5$ \\
        NEN - SPSA & $ 16.75 \pm \scriptstyle 1.64$ & $50.13 \pm \scriptstyle 4.16$ \\
        NFN(\textit{Fixed}) - AD & $ 2.37 \pm \scriptstyle 0.04$ & $-$ \\
        NFN(\textit{Fixed}) - BOND & $ 3.92 \pm \scriptstyle 0.01$ & $99.95 \pm \scriptstyle 0.02$ \\
        NFN(\textit{Fixed}) - BONDS & $ 3.93 \pm \scriptstyle 0.02$ & $76.13 \pm \scriptstyle 22.94$ \\
        NFN(\textit{Fixed}) - SPSA & $ 3.03 \pm \scriptstyle 0.04$ & $50.03 \pm \scriptstyle 8.44$ \\
        \bottomrule
    \end{tabular}
    
\end{table}

\subsection{Analysis of Perturbation Magnitudes} \label{appendix:perturbation_magnitudes}
\begin{figure}[htbp]
    \centering
    \includegraphics[width=0.85\textwidth]{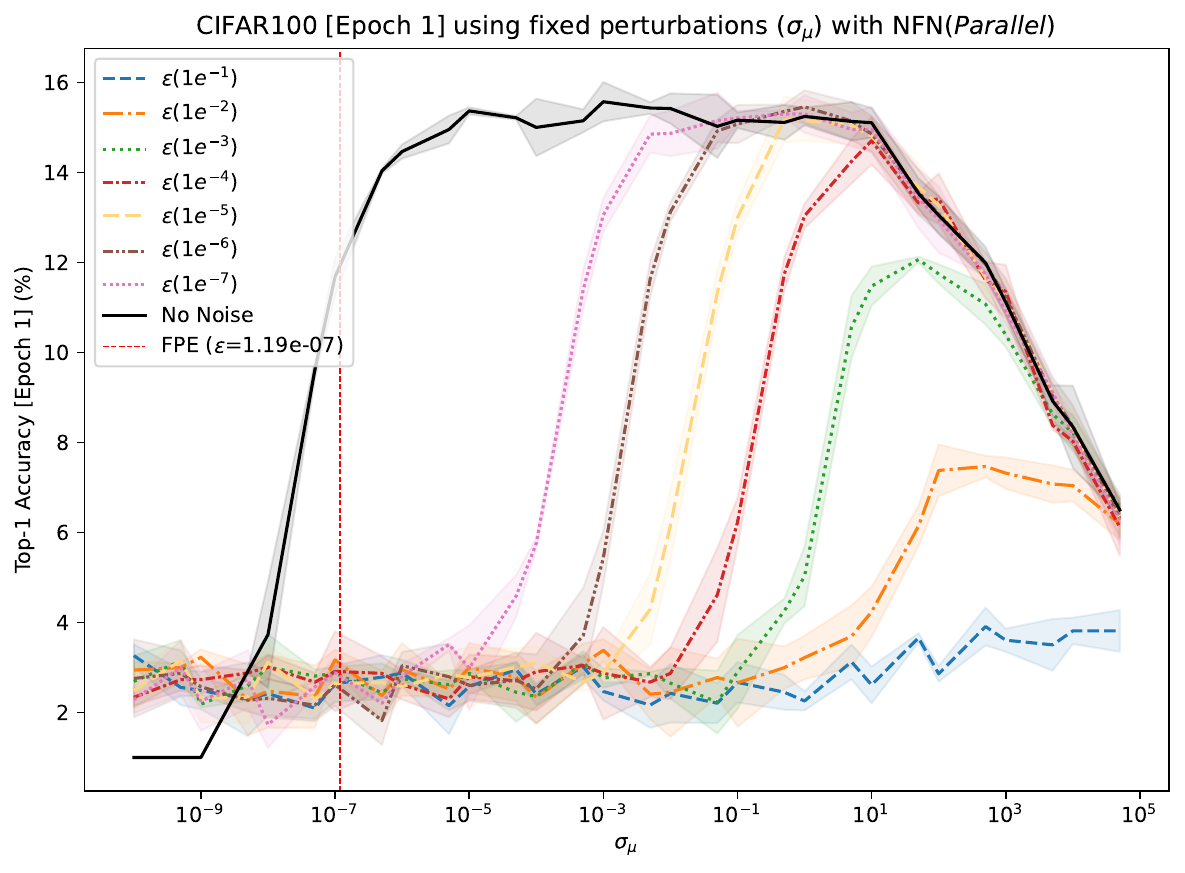}
    \caption{Top 1\% accuracy at epoch 1 for NFN(\textit{Parallel}) models with fixed perturbations magnitudes. We also show models where noise is added to reservoir outputs $\varepsilon(X)$ where $\varepsilon \sim \mathcal{U}[0,X]$} 
    \label{fig:fixed_perturbation_sweep}
\end{figure}

We seek to analyze the role of perturbation magnitude in achieving optimal convergence with ZO methods. To do this, we implement a version of BOND which uses a constant smoothing coefficient $\sigma_{\mu}$, sampling perturbations $\Delta_{x}^t$ from the normal distribution $\mathcal{N}(0,\sigma_{\mu})$ (similar to FDSA and SPSA variants). For the smooth function $f(x)\in C^{\infty}$, taking $\sigma_{\mu}\rightarrow 0$ facilitates a reliable gradient estimate, effectively acting as a limit for the numerical approximation of the gradient. However, in practice, most functions possess the quality of smoothness only above a particular scale $f \in C^\infty([a,b]) \quad \text{for all } [a,b] \subseteq \mathbb{R} \text{ with } b - a \geq \delta_{min}$. Consider the example of classical computation, where functions inevitably suffer from non-smoothness due to floating point errors at the very small scale. At these scales, numerical gradient approximations tend to falter, leading to suboptimal performance. We illustrate this effect in Figure \ref{fig:fixed_perturbation_sweep}, where we sweep $\sigma_{\mu} \in [1e^{-10}, 1e^{5}]$, and observe that performance degrades when $\sigma_{\mu}\rightarrow 0$. This highlights the importance of a method such as BOND which determines optimal perturbation bounds, rather than selecting a fixed or scheduled hyperparameter (as in existing ZO variants). In Figure~\ref{fig:fixed_perturbation_sweep}, we also provide implementations where we add uniformly distributed noise $\varepsilon(X)$ to the reservoir outputs, highlighting potential considerations for a noisy physical system. We use 32 bit operations, where machine error is given by $\epsilon = 1.19e^{-7}$.

With Figure~\ref{fig:fixed_perturbation_sweep}, we contend that the generally uniform performance across $10^{-6}\leq \sigma_{\mu} \leq 10^{1}$ for \enquote{No Noise} indicates that sign accuracy, rather than outright gradient estimation precision, drives convergence. If convergence depended solely on estimation error, we would expect a performance peak rather than the observed plateau. Additionally, we observe that cases where the added noise is $1e^{-5}$ or less are capable of achieving equivalent performance to a \enquote{No Noise} system -- albeit with more constrained requirements on optimal perturbation magnitudes. For cases where the noise upper bound exceeds $1e^{-5}$, there is an increasing reduction in maximum performance. Given the specificity of our implementation, any claims regarding the explicit noise threshold required to enable optimality would likely not generalize. Rather, Figure~\ref{fig:fixed_perturbation_sweep} is useful because it highlights potential for the implementation of physical reservoir systems, where achieving near-zero noise may be highly-complex, but achieving some sub-threshold (i.e. around $1e-5$) noise may be attainable.

\begin{figure}[htbp]
    \centering
    \includegraphics[width=0.75\textwidth]{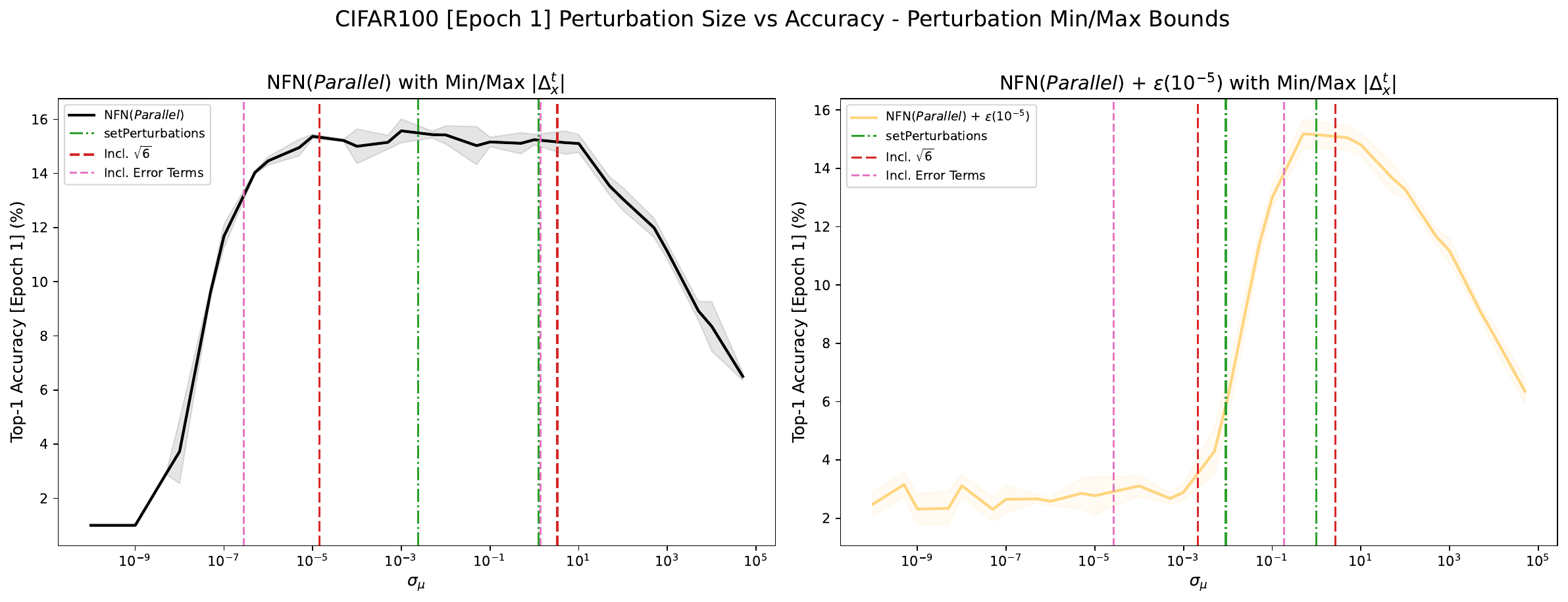}
    \vspace{0.5cm}
    \includegraphics[width=0.75\textwidth]{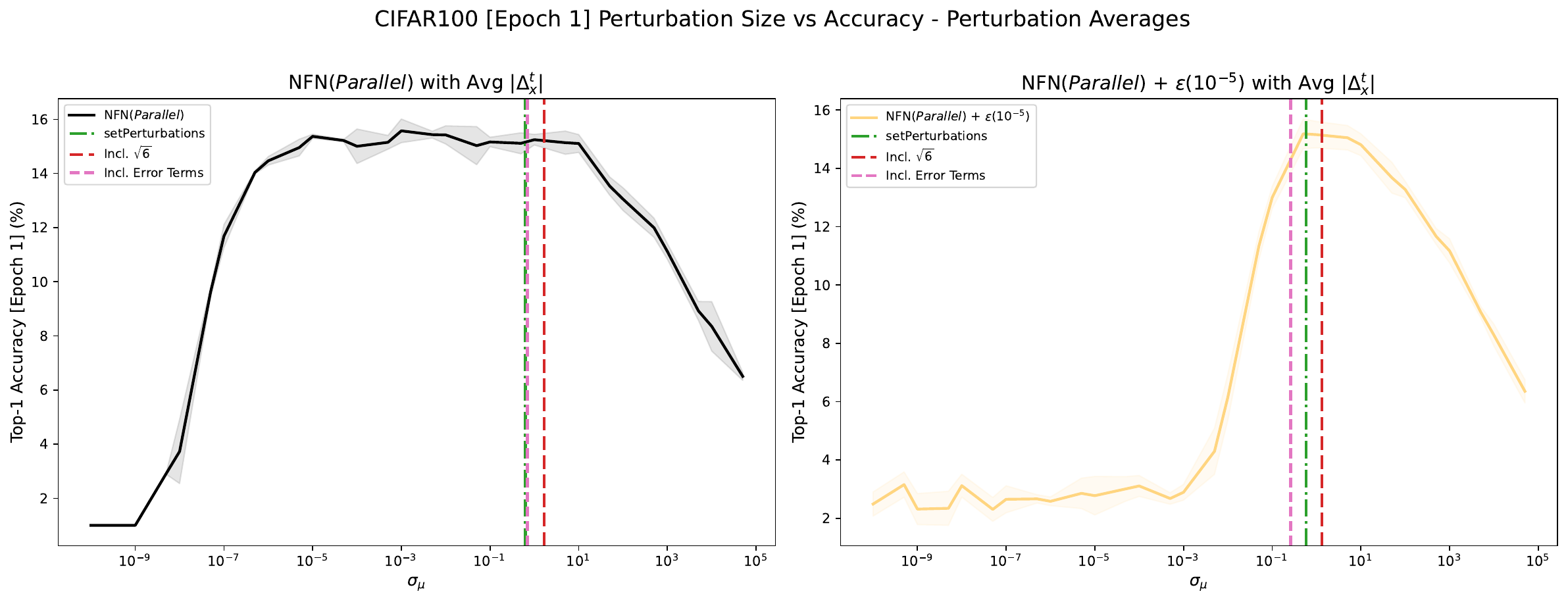}
    \caption{Top: min/max bounds of each formulation for the \enquote{No Noise} and $\varepsilon(1e^{-5})$ implementations of NFN(\textit{Parallel}). Bottom: Equivalent to Top row, demonstrating avg perturbation magnitudes instead of min/max bounds.}
    \label{fig:stacked}
\end{figure}

As increasing noise reduces the window of optimal perturbation magnitudes, we seek to verify that BOND effectively adjusts the perturbation bounds to ensure maximum performance capabilities. We illustrate the min/max bounds and average perturbation magnitudes of different BOND formulations in Figure~\ref{fig:stacked}, using Table~\ref{tab:bound_testing} as a reference.
\begin{table}[htbp]
    \caption{Formulations used to demonstrate perturbation magnitudes in Figure~\ref{fig:stacked}}
    \label{tab:bound_testing}
    \centering
    \begin{tabular}{l c c}
        \toprule
        \textbf{Label} & \textbf{Lower Bound ($\underline{\Delta_x^t}$)} & \textbf{Upper Bound ($\overline{\Delta_x^t}$)} \\
        \midrule
        setPerturbations & $\displaystyle \frac{b}{\min_{i\in d_{\mathcal{R}}} \{\lvert\hat{m}^{\mathcal{R}}_{t-1}\rvert\}}$ & $\displaystyle \alpha_{t} \cdot \sqrt{\frac{\lvert\hat{m}^{\mathcal{R}}_{t-1}\rvert}{\hat{v}^{\mathcal{R}}_{t-1}}}$ \\[2em]
        Incl. $\sqrt{6}$ & $\displaystyle \frac{b}{\lvert \hat{m}^{\mathcal{R}}_{t-1} \rvert+\epsilon}$ & $\displaystyle \alpha_{t} \cdot \sqrt{\frac{6\lvert\hat{m}^{\mathcal{R}}_{t-1}\rvert}{\hat{v}^{\mathcal{R}}_{t-1}+\epsilon}}$ \\[2em]
        Incl. Error Terms & Equation~\ref{eq:lower_bound_full} & Equation~\ref{eq:upper_bound_full} \\[0.5em]
        \bottomrule
    \end{tabular}
\end{table}

\begin{figure}[htbp]
    \centering
    \begin{minipage}{0.48\textwidth}
        \centering
        \textbf{(a)} \\[0.2cm]
        \includegraphics[width=\textwidth]{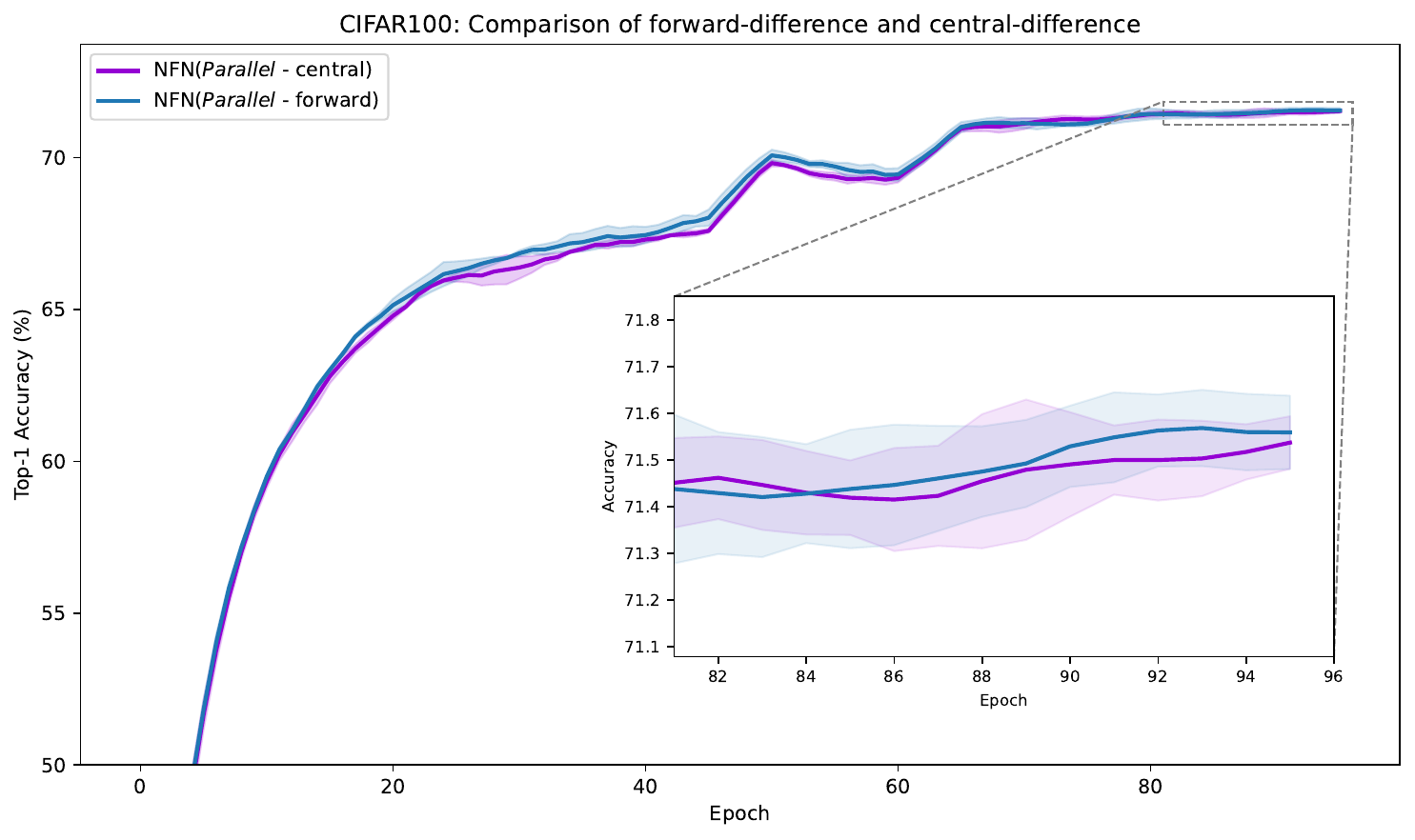}
    \end{minipage}
    \hfill
    \begin{minipage}{0.48\textwidth}
        \centering
        \textbf{(b)} \\[0.2cm]
        \includegraphics[width=\textwidth]{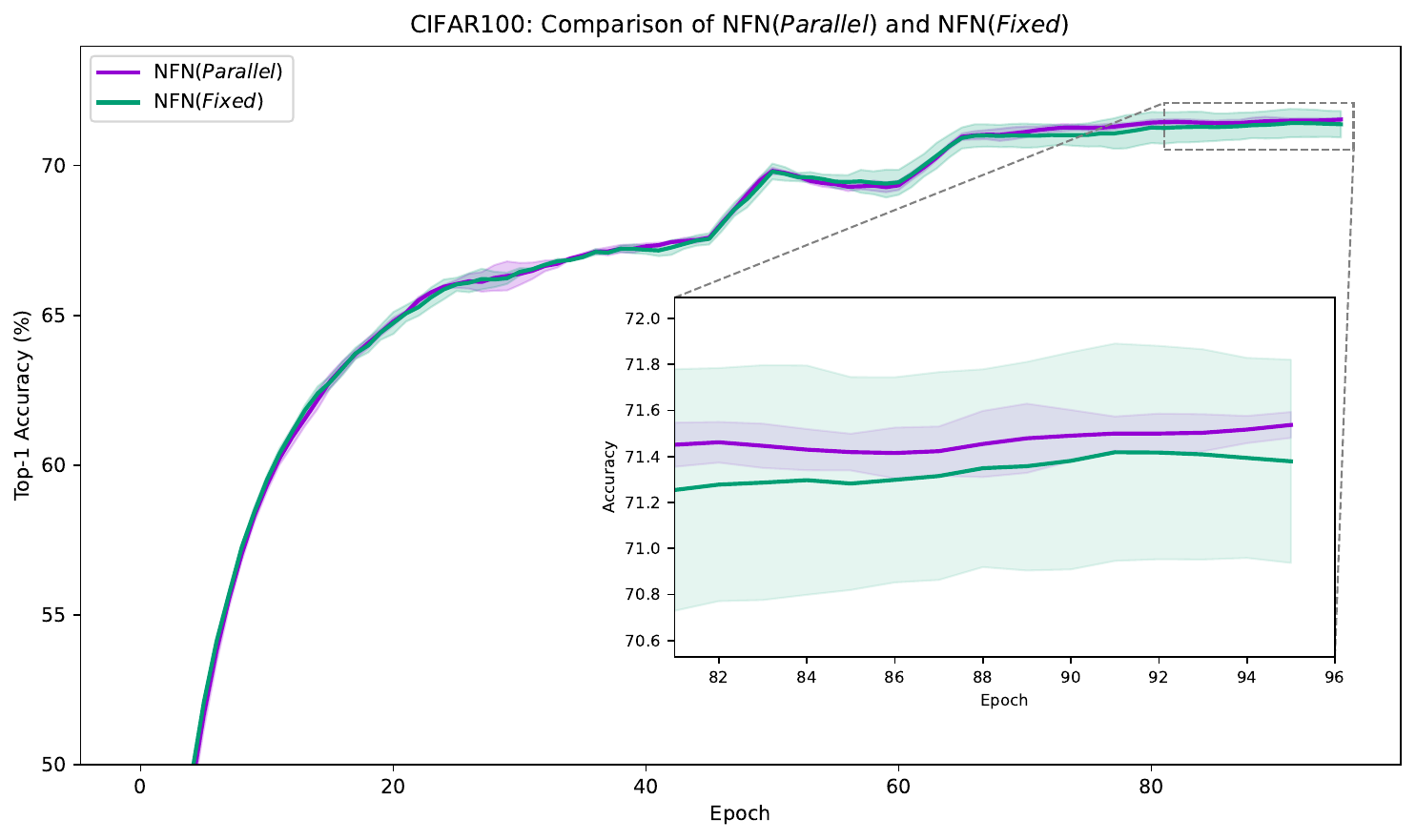}
    \end{minipage}
    \vspace{0.4cm}
    \begin{minipage}{0.48\textwidth}
        \centering
        \textbf{(c)} \\[0.2cm]
        \includegraphics[width=\textwidth]{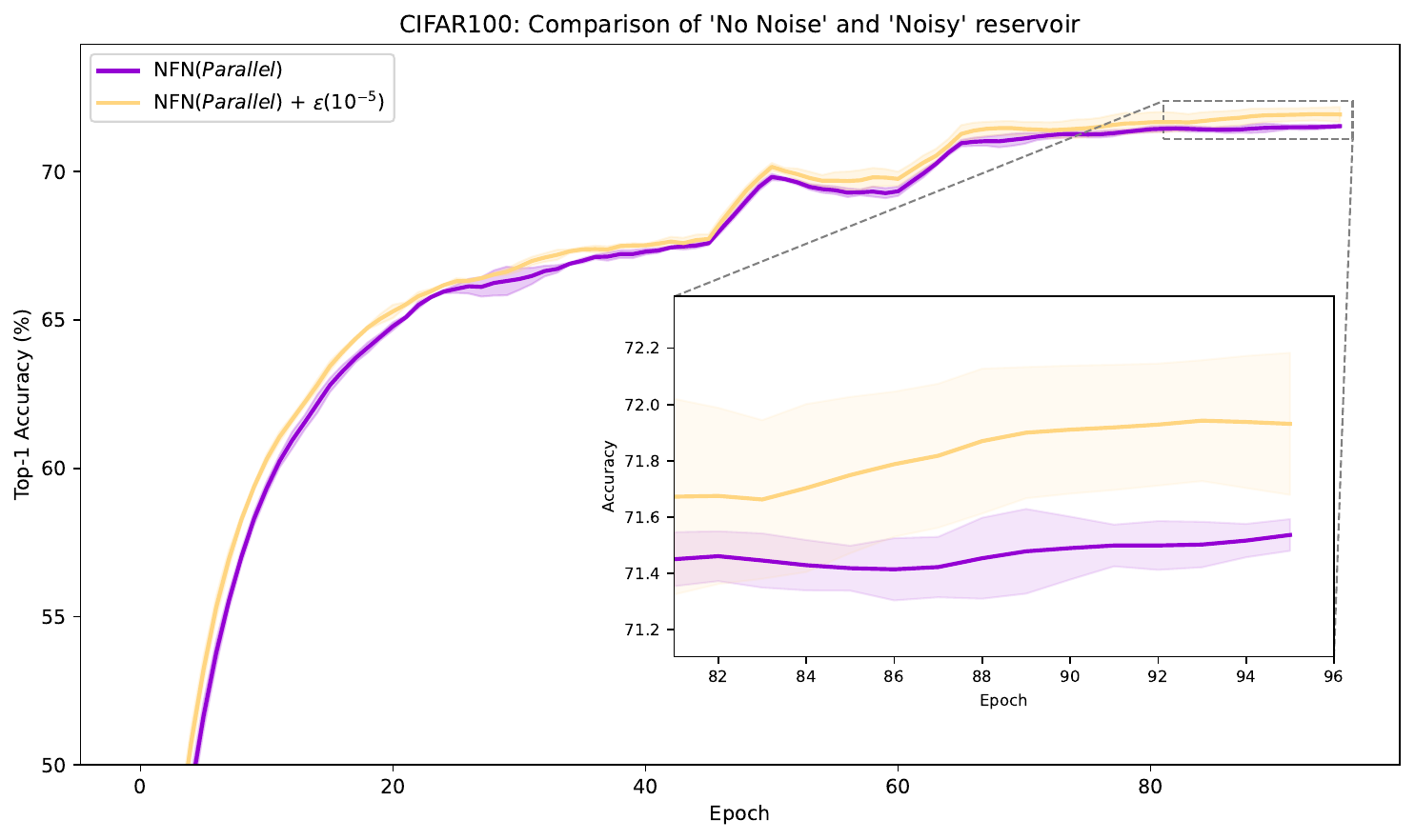}
    \end{minipage}
    \hfill
    \begin{minipage}{0.48\textwidth}
        \centering
        \textbf{(d)} \\[0.2cm]
        \includegraphics[width=\textwidth]{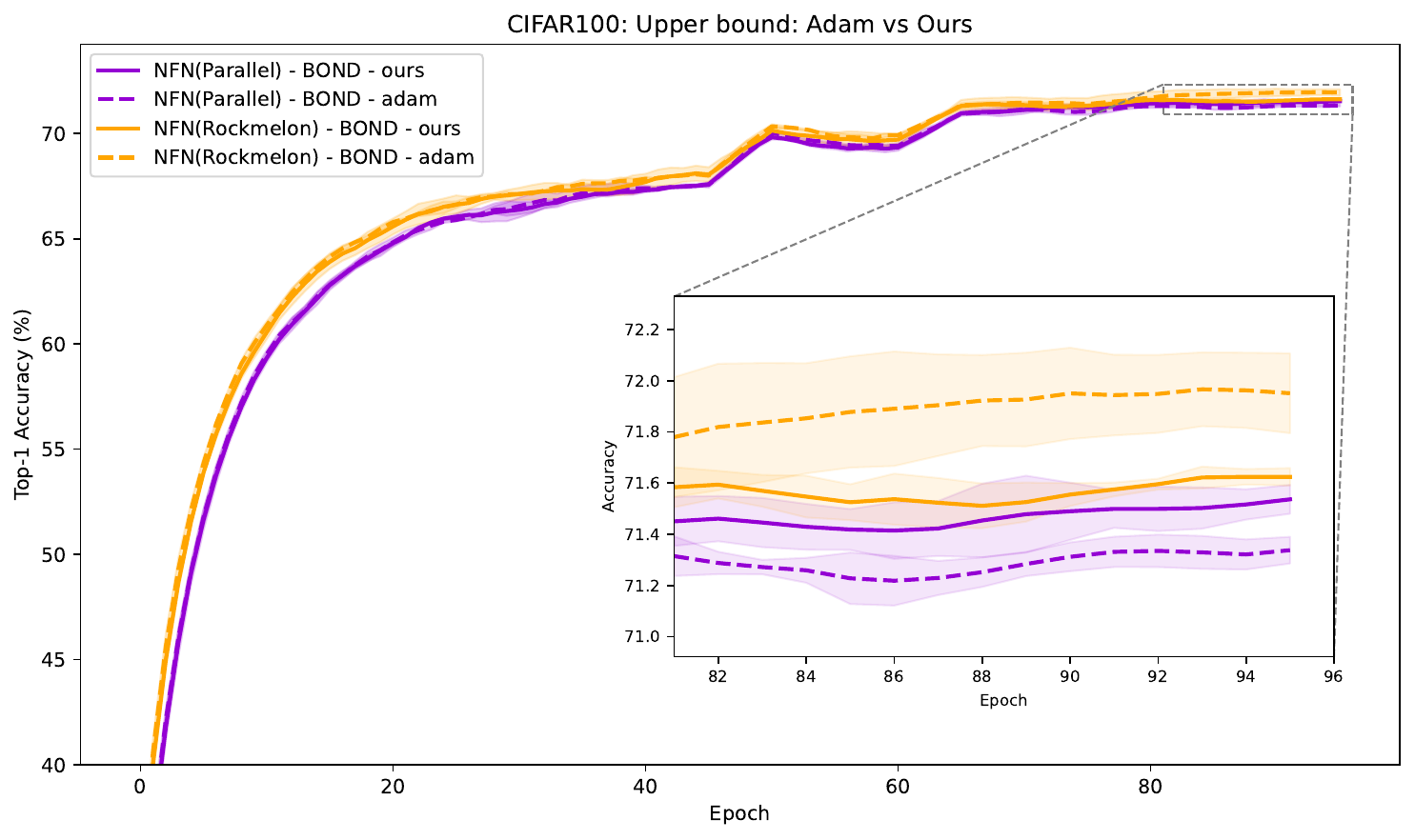}
    \end{minipage}
    \caption{(a) Forward-difference vs Central-difference. (b) Parallel vs Fixed structure. (c) No noise vs Noisy Reservoir. (d) Adam as an upper bound vs BOND.}
    \label{fig:four_plots}
\end{figure}

Figure~\ref{fig:stacked} illustrates that Equation~\ref{eq:lower_bound_used} and \ref{eq:upper_bound_used} from \enquote{setPerturbations} do indeed determine desirable perturbation magnitudes, even when noise is added. The formulation \enquote{Incl. $\sqrt{6}$ } leads to similar bound placements, however we contend that the slightly larger bounds result in increased variability in the general performance - a phenomenon observed throughout initial testing and discussed further with Figure~\ref{fig:four_plots}. We also note that the inclusion of error terms in \enquote{Incl. Error Terms} reduces the upper bound which affects overall performance. As $\underline{\Delta_x^t}$ tends to be orders of magnitude smaller than $\overline{\Delta_x^t}$, sampling uniformly on a linear-scale tends to result in biased log-scale values towards $\overline{\Delta_x^t}$, which we observe to be a beneficial property. 

\begin{figure}[htbp]
    \centering
    \includegraphics[width=0.85\textwidth]{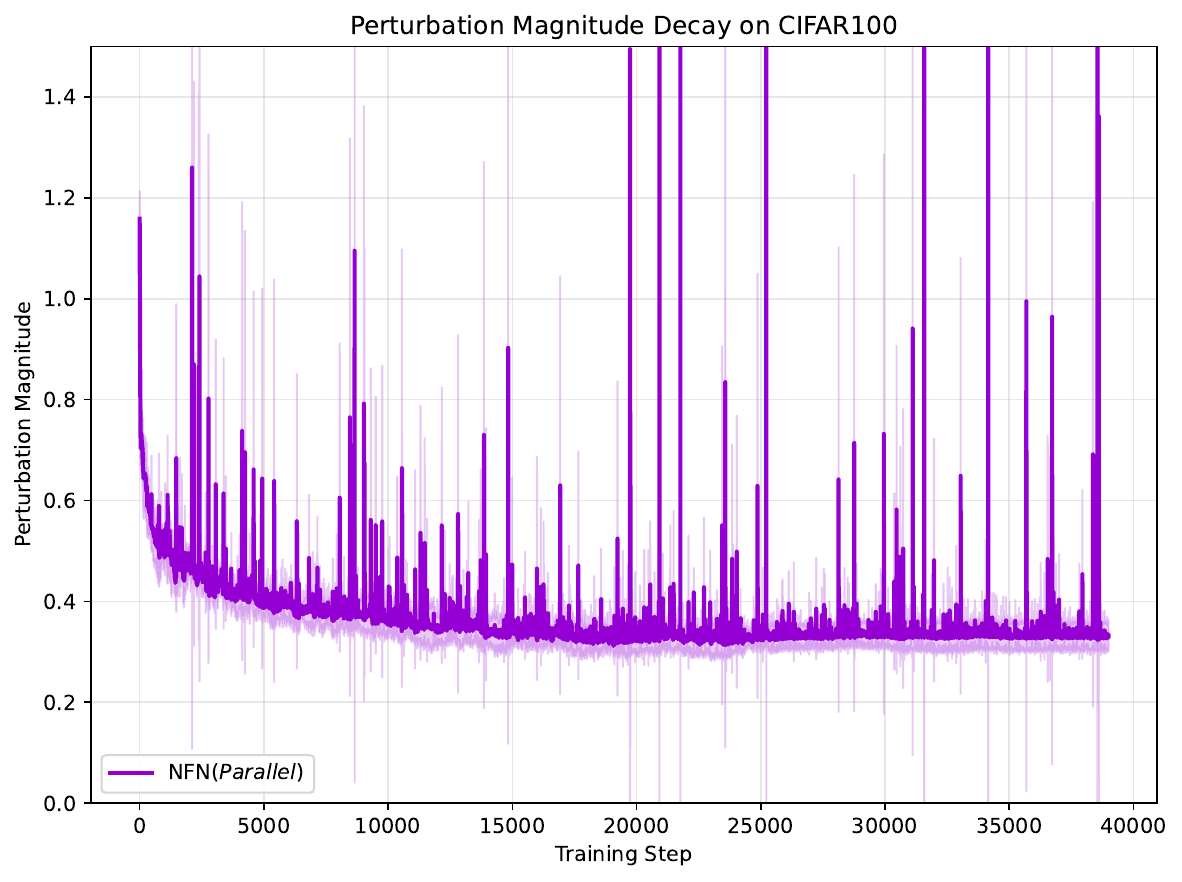}
    \caption{Decay of perturbation magnitudes across training.}
    \label{fig:decaying_perturbation_magnitudes}
\end{figure}

We provide a plot of how the perturbation magnitudes of \enquote{setPerturbations} evolve throughout training in Figure~\ref{fig:decaying_perturbation_magnitudes}. The perturbation magnitudes exhibit relatively stable dynamics, on average decaying comparably to the gradient norm. Naturally, this effect enables us to bypass existing ZO complexities such as the rate of scheduling for smoothing coefficients. An interesting note is the variability of the perturbation magnitudes, with some large spikes occurring, particularly in late stage training. In initial testing we had a rescaling component which normalized gradient estimates and therefore also regularized perturbation magnitudes. Ultimately we observed that this added computation, and actually often led to a slight reduction in performance. For physical systems where the input values may be constrained, we note that clamping the BOND perturbation magnitudes may be necessary.

A through-line within the NFN($\textit{Parallel}$) CIFAR100 experimentation is the utility of (a limited amount of) noise. In Figure~\ref{fig:four_plots}c we show the full BOND training comparison of a reservoir with and without added noise, observing that the $\varepsilon(1e^{-5})$ noise actually improves performance. The added noise in gradient estimation induced by forward-difference compared to central-difference approximations led to comparable results in Figure~\ref{fig:four_plots}a, despite central-difference having improved sign accuracy in Table~\ref{table:forwardvscentral}. An interesting extension of this discussion is captured in Figure~\ref{fig:four_plots}d, where we plot BOND convergence using the conventional \enquote{Adam} formula $\hat{m}/\sqrt{\hat{v}}$ as the upper bound formulation, compared to our implementation $\sqrt{\hat{m}/\hat{v}}$. As noted in Table~\ref{table:forwardvscentral} and Table~\ref{table:bondvsadam_sign}, our method attains slightly better sign estimates and performance. The only exception observed is that an \enquote{Adam} upper bound does better for NFN(\textit{Rockmelon}) despite having slightly worse sign-accuracy. In this light, we suggest that Adam (as an optimizer) may not explicitly be the most precise estimator of step size, but rather accurate enough that the noise induced by its error is beneficial, particularly so for sharp loss landscapes. This is certainly not a new concept and is thematically similar to how noise through mini-batching can be beneficial \citep{noisy_sgd_2020,noisy_sgd_2022}. However, we find it interesting perk of BOND, which may be useful in further work. We suggest that the utility of noise also strengthens the argument for the plausibility of using physical reservoir systems with sub-threshold noise as functional transforms. Lastly, we provide a comparison of the NFN($\textit{Fixed}$) and NFN($\textit{Parallel}$) architectures in Figure~\ref{fig:four_plots}b. Demonstrating that the modularized, Parallel architecture tends to perform more consistently.

\begin{table}[h]
    \caption{Training Information - Difference}
    \label{table:forwardvscentral}
    \centering
    \begin{adjustbox}{width=0.6\textwidth}
    \begin{tabular}{lrr}
        \toprule
        Model  &Loop Time (ms)& Correct Sign (\%) \\
        \midrule
        NFN(\textit{Parallel}) & $ 389.07 \pm \scriptstyle 0.11$ & $99.85 \pm \scriptstyle 0.01$ \\
        NFN(\textit{Fixed}) & $ 280.39 \pm \scriptstyle 0.01$ & $99.99 \pm \scriptstyle 0.01$ \\
        NFN(\textit{Parallel}) - forward-diff & $ 318.57 \pm \scriptstyle 0.01$ & $98.53 \pm \scriptstyle 0.03$ \\
        NFN(\textit{Parallel}) + $\varepsilon(1e^{-5})$ & $ 392.32 \pm \scriptstyle 0.03$ & $99.54 \pm \scriptstyle 0.01$ \\
        NFN(\textit{Parallel}) - upper=adam & $ 394.53 \pm \scriptstyle 0.02$ & $99.34 \pm \scriptstyle 0.07$ \\
        NFN(\textit{Rockmelon}) & $ 1566.91 \pm \scriptstyle 15.21$ & $99.90 \pm \scriptstyle 0.04$ \\
        NFN(\textit{Rockmelon}) - upper=adam & $ 1547.79 \pm \scriptstyle 219.52$ & $99.82 \pm \scriptstyle 0.07$ \\
        \bottomrule
    \end{tabular}
    \end{adjustbox}
\end{table}

\begin{table}[ht]
    \caption{BOND upper bound vs Adam upper bound.}
    \label{table:bondvsadam_sign}
    \centering
    \begin{adjustbox}{width=\textwidth}
    \begin{tabular}{lrrrrrr}
        \toprule
        & \multicolumn{2}{c}{FashionMNIST} & \multicolumn{2}{c}{CIFAR10} & \multicolumn{2}{c}{CIFAR100} \\
        \cmidrule(lr){2-3} \cmidrule(lr){4-5} \cmidrule(lr){6-7}
        Model & \shortstack{Acc. \\ (\%)} & \shortstack{Sign Acc. \\ (\%)} & \shortstack{Acc. \\ (\%)} & \shortstack{Sign Acc. \\ (\%)} & \shortstack{Acc. \\ (\%)} & \shortstack{Sign Acc. \\ (\%)} \\
        \midrule
        BOND - Adam & $93.40_{\pm \scriptstyle 0.07}$ & $99.869_{\pm \scriptstyle 0.023}$ & $93.10_{\pm \scriptstyle 0.06}$ & $99.978_{\pm \scriptstyle 0.004}$ & $71.30_{\pm \scriptstyle 0.24}$ & $99.989_{\pm \scriptstyle 0.001}$ \\
        BOND - Ours & $\textbf{93.42}_{\pm \scriptstyle 0.10}$ & $\textbf{99.907}_{\pm \scriptstyle 0.013}$ & $\textbf{93.25}_{\pm \scriptstyle 0.18}$ & $\textbf{99.994}_{\pm \scriptstyle 0.001}$ & $\textbf{71.32}_{\pm \scriptstyle 0.49}$ & $\textbf{99.995}_{\pm \scriptstyle 0.001}$ \\
        \bottomrule
    \end{tabular}
    \end{adjustbox}
\end{table}

\newpage
\subsection{Simultaneous methods: BONDS and SPSA} \label{appendix:bondsvsspsa}
\begin{figure}[H]
    \centering
    \includegraphics[width=\textwidth]{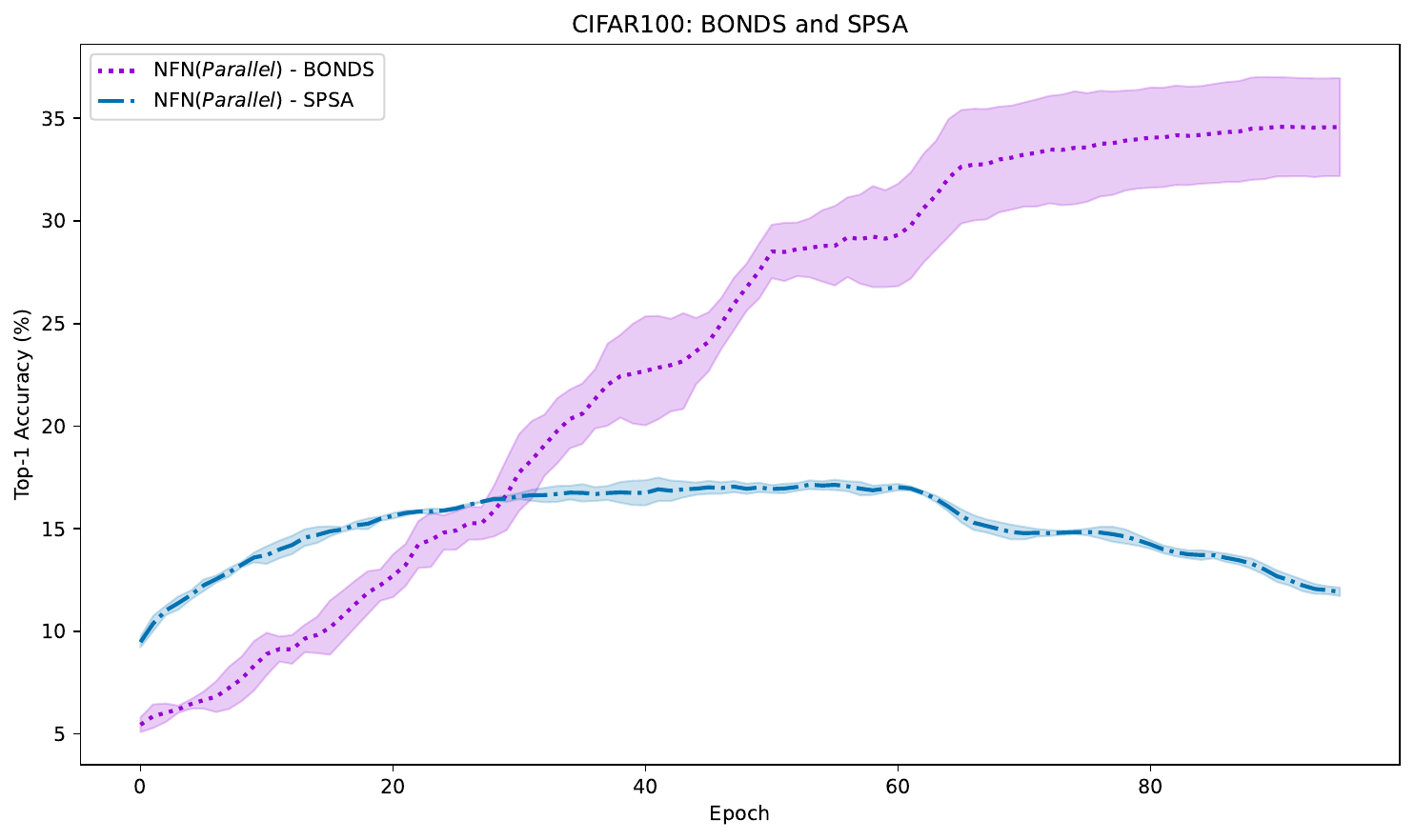}
    \caption{The performance of NFN(\textit{Parallel}) with SPSA and BONDS on CIFAR100}
    \label{fig:bondsVSspsa}
\end{figure}

In Figure \ref{fig:bondsVSspsa}, we show the performance of BONDS and SPSA for CIFAR100. It remains clear that simultaneous methods, BONDS and SPSA, are not comparable to AD or BOND, especially as input dimensionality increases. 

For SPSA, we found that SGD meaningfully outperformed Adam optimization, noted in the experimental set up, see Appendix~\ref{appendix:setup_cifar100}. SPSA also required a significantly smaller learning rate with SGD, using $\eta=1e-6$ to achieve optimal performance. We note that only NFN(\textit{Fixed}) performed worse than NFN(\textit{Parallel}) for BONDS and SPSA, leading to their omission for readability.

\subsection{Reservoir Scaling} \label{appendix:reservoir_scaling}
The overfitting observed during \textit{Rockmelon} training indicates the need for tailored optimization and learning rate schedules to better accommodate its functional complexity. The performance of NFN(\textit{Parallel}) and NFN(\textit{Rockmelon}) supports reservoir modularization as a potential method for improving the viability of hardware applications. Overall, these findings corroborate existing observations from RP methods~\cite{rand_projections} while extending them in two key ways: demonstrating compatibility with gradient estimation, and showing that structured reservoirs outperform unstructured ones.

For readability, we omit NFN(\textit{Fixed}) and simultaneous perturbation methods (SPSA and BONDS) from Figure~\ref{fig:cifar100} --- these results appear in Appendix~\ref{appendix:perturbation_magnitudes}, Figures~\ref{fig:fixed_reservoir} and~\ref{fig:bondsVSspsa}. We note that NFN(\textit{Fixed}) is comparable to NFN(\textit{Parallel}), whilst both SPSA and BONDS achieve poor convergence.

In initial experiments, we explored alternative black-box functions including orthogonalized kernel transforms, Hamiltonian evolutions, polynomial functions and modelled mechanical systems --- generally yielding variable results but not exceeding the performances of Figure~\ref{fig:reservoir_scaling} and \ref{fig:cifar100}. In this light, we recognize that additional work is required to develop a comprehensive understanding of the useful properties of a black-box function.

\begin{table}[ht]
    \caption{CIFAR100 Results}
    \label{table:cifar100}
    \centering
    \begin{adjustbox}{width=0.7\textwidth}
    \begin{tabular}{lrrrr}
        \toprule
        Model & \shortstack{Acc. \\ (\%)} & \shortstack{Sign Acc. \\ (\%)} & \shortstack{Loop Time \\ (ms)} & \shortstack{Gain \\ (\%)} \\        
        \midrule
        No Reservoir - AD & $71.09_{\pm \scriptstyle 0.14}$ & $-$ & $259.47_{\pm \scriptstyle 0.19}$ & $-$ \\
        NFN(\textit{Fixed}) - AD & $71.1_{\pm \scriptstyle 0.12}$ & $-$ & $247.88_{\pm \scriptstyle 0.0}$ & \textcolor{green!70!black}{$\uparrow +0.01$} \\
        NFN(\textit{Parallel}) - AD & $71.5_{\pm \scriptstyle 0.25}$ & $-$ & $251.33_{\pm \scriptstyle 0.03}$ & \textcolor{green!70!black}{$\uparrow +0.41$} \\
        NFN(\textit{Rockmelon}) - AD & $\textbf{72.2}_{\pm \scriptstyle 0.21}$ & $-$ & $279.11_{\pm \scriptstyle 0.65}$ & \textcolor{green!70!black}{$\uparrow +1.11$} \\
        NFN(\textit{Fixed}) - BOND & $71.32_{\pm \scriptstyle 0.49}$ & $99.99_{\pm \scriptstyle 0.01}$ & $280.33_{\pm \scriptstyle 0.05}$ & \textcolor{green!70!black}{$\uparrow +0.23$} \\
        NFN(\textit{Parallel}) - BOND & $71.58_{\pm \scriptstyle 0.07}$ & $99.88_{\pm \scriptstyle 0.02}$ & $389.02_{\pm \scriptstyle 0.05}$ & \textcolor{green!70!black}{$\uparrow +0.49$} \\
        NFN(\textit{Rockmelon}) - BOND & $\textbf{71.62}_{\pm \scriptstyle 0.07}$ & $99.93_{\pm \scriptstyle 0.01}$ & $1568.62_{\pm \scriptstyle 2.91}$ & \textcolor{green!70!black}{$\uparrow +0.53$} \\
        \bottomrule
    \end{tabular}
    \end{adjustbox}
\end{table}

Table~\ref{table:cifar100} establishes that performance gains for NFN(\textit{Parallel}) and NFN(\textit{Rockmelon}) are attributable to the reservoir inclusion rather than just auxiliary effects like noise injection, as both models exhibit substantial improvements with both AD and BOND. The less structured \textit{Fixed} reservoir does not exhibit equivalent improvements under AD, potentially indicating that it does not possess the same advantageous properties as \textit{Parallel} and \textit{Rockmelon}.

We also note that the relative latency of BOND can be small, only 13\% slower than AD for NFN(\textit{Fixed}). However, we underscore the importance of the relative bottleneck of reservoir run-time: the larger NFN(\textit{Rockmelon}) ($|\theta_{Rockmelon}|>>|\theta_{Fixed}|$) incurs a 462\% slowdown relative to equivalent AD. This has important ramifications for potential hardware design and implementations.

\newpage
\section{Reservoir Configurations} \label{appendix:reservoir_functions}
In this section, we provide information regarding the configurations used for implementing black-box functions as neural network architectures. We also discuss a \textit{Parallel} architecture, which decomposes the reservoir into modules that each act on a unique subset of the reservoir inputs. This contrasts the \textit{Fixed} design which is fully connected and does not permit concurrent perturbations. This approach can further mitigate the effect of estimation complexity.

\subsection{Echo-State-Network}
ESN's are particularly well suited to time series problems \citep{LUKOSEVICIUS2009127,jaeger2}. Whilst we do not implement a time series problem in our work, we contend that using an ESN provides important information on the numerical differentiability of recurrent and noisy systems. This is an important consideration for future work, especially in the context of recent explorations on the utility of inherent temporal dynamics in physical RC devices for time series applications \citep{hsbc,quantum_spin_timeseries}.

\subsection{\textit{Fixed} Reservoir}
In our experimentation, a \textit{Fixed} reservoir is a single, fully connected, randomly-initialized FFNN whose weights are frozen during training. We implement Tanh activations to preserve smoothness. This architecture allows for the dual comparison of recursive vs frozen, and fully-connected vs modularized functions. We show a diagram of the \textit{Fixed} reservoir in Figure~\ref{fig:fixed_reservoir}.

\subsection{\textit{Parallel} Reservoir}
\begin{figure}[htbp]
    \centering
    \begin{minipage}{0.5\textwidth}
        \centering
        \includegraphics[width=\textwidth]{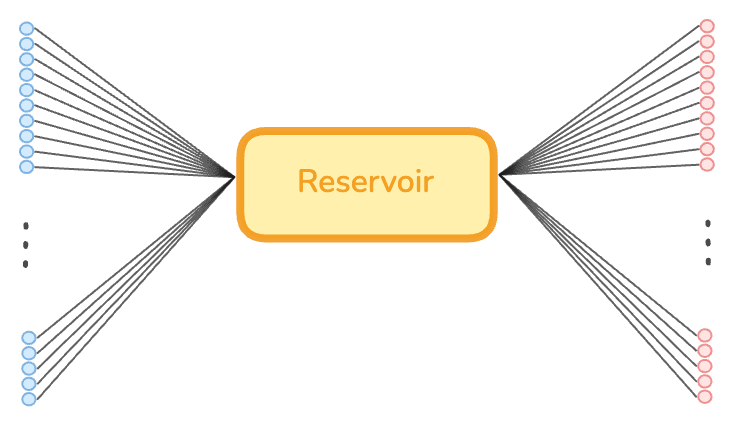}
        \caption{\textit{Fixed} Reservoir}
        \label{fig:fixed_reservoir}
    \end{minipage}%
    \hfill
    \begin{minipage}{0.5\textwidth}
        \centering
        \includegraphics[width=\textwidth]{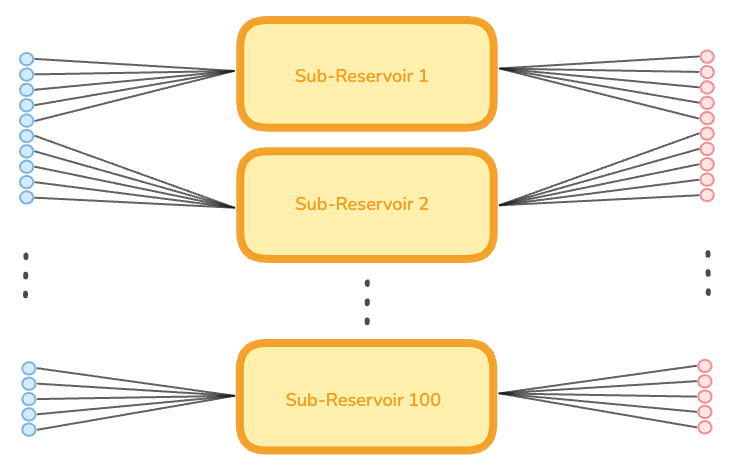}
        \caption{\textit{Parallel} Reservoir}
        \label{fig:parallel_reservoir}
    \end{minipage}
\end{figure}

The \textit{Parallel} reservoir consists of multiple independent FFNN's (sub-reservoirs). Each sub-reservoir processes a distinct subset of the inputs separately and the outputs from all are concatenated to form the total reservoir output. We illustrate this structure in Figure~\ref{fig:parallel_reservoir}. The \textit{Parallel} reservoir enables perturbation parallelization, where each sub-reservoir can execute BOND type methods independently. This attribute may facilitate improvements in the computational bottleneck of BOND in future applications. Additionally, it introduces the notion that physical reservoirs can be modular, rather than having to be a single, fully integrated device. This hints towards approaches where the combined functionality of the many devices can be used to design architectures or black-box functions with specific functions or properties.

\newpage
\section{Experimental Setups}
In this section, we provide all details about the experiments used in the paper. We note that for SPSA and BONDS implementations, the hyperparameter $n_{\textit{pert}}=1$ was used throughout. We did experiment with increasing this value (using $n_{\textit{pert}}=20$) but obtained non-meaningful performance gains for the increased computational cost. We do not include FDSA methods in this work, as the size of our \textit{read-in} networks ($d_{\theta_A}$) prohibits relevant experimentation. Naturally we expect that a well-tuned FDSA approach would have somewhat similar performance to BOND. This extends to the application of BOND-type methods using fixed perturbation sizing as depicted in Figure~\ref{fig:fixed_perturbation_sweep}, where any carefully selected perturbation regime should be approximately equivalent to BOND and AD. In this sense, the bounds of BOND are valuable as they eliminate the process required for careful selection, which can be particularly non-trivial on physical RC devices (we observed high variability of noise floor when working on quantum devices, Appendix~\ref{appendix:setup_rockmelon}). \\

For BOND, we select the parameter $b=1.19e^{-7}$ to initialize perturbation magnitudes at iteration $t=0$. This value is based off the machine level precision at fp32, and we find it performs relatively well. We note that significantly poor estimates of $b$ can lead to poor convergence. For physical systems, we expect that this bound for the baseline noise distribution is generally observable and can be known prior to training.\\

In experimentation, we compare the approach of retaining and overwriting a single computational graph against maintaining two separate computational graphs for $f_A(\cdot,\,\theta_{A})$ and $f_B(\cdot,\,\theta_{B})$. This allows for the comparison of BOND and AD gradients, and ensures no gradient leakage. Note, all results correspond to performance at final epoch, as using \enquote{best epoch} (in our observations) tended to slightly biased the results towards BOND methods, primarily to due slightly increased performance variability likely associated with gradient noise.

\subsection{Sign Accuracy and Convergence - CHP}\label{appendix:setup_sign_accuracy_and_convergence}
\textsc{[Pretains to Figure~\ref{fig:calhouse_loss}, and Table~\ref{table:calhouse}]}

For CHP data, we scale target values, dividing by $10,000$. We use a learning rate scheduler with an initial learning rate of $1e-3$ for all models, except SPSA which benefited from a larger learning rate at $1e-2$. The details of the lambda scheduling coefficients are shown below where $e$ is the current epoch and $E$ is the total number of epochs.
\[
\alpha_e = \begin{cases}
1.0 & \text{if } e < 0.5E \\
0.1 & \text{if } 0.5E \leq e \leq 0.75E \\
0.01 & \text{if } e > 0.75E
\end{cases}
\]
Additional details include.

{\leftskip=2em
\noindent Loss Function: Huber.\newline
Optimizer: Adam ($\beta_1=0.9$, $\beta_2=0.999$, $\textit{weight\_decay}=0$ ), Epochs: 100. Batch size: 64. Seeds: [42, 63, 14]. GPU: A10G.

\subtitleblock{Read-in Network ($f_A$) - FFNN}{
    Input/Output Dim: 8/5. Hidden Layers: [100,100]. Activation: ReLU.}
\subtitleblock{Black-Box Function [NEN] ($\mathcal{R}$) - ESN}{
    Input/Output Dim: 5/5. Hidden State: [200]. Activation: Tanh. Row Normalization: True}
\subtitleblock{Black-Box Function [NFN] ($\mathcal{R}$) - FFNN}{
    Input/Output Dim: 5/5. Hidden Layers: [500]. Activation: Tanh.}
\subtitleblock{Read-out Network ($f_B$) - FFNN}{
    Input/Output Dim: 5/1. Hidden Layers: [100,100]. Activation: Tanh.}
}

\subsection{Sign Accuracy and Convergence - FashionMNIST and CIFAR10} \label{appendix:fmnist_cifar10_setup}
\textsc{[Pretains to Tables~\ref{table:fmnist_cifar10},\ref{table:bondvsadam_sign}]}

These experiments are intended to support findings in CHP experiments, demonstrating that BOND achieves good sign accuracy, and AD-equivalent convergence. We adopt a new learning rate scheduler, with an initial learning rate of $1e-3$.
\[
\alpha_e = \begin{cases}
1.0 & \text{if } e < 0.5E \\
0.5 & \text{if } 0.5E \leq e \leq 0.65E \\
0.1 & \text{if } 0.65E \leq e \leq 0.8E \\
0.05 & \text{if } 0.8E \leq e \leq 0.9E \\
0.01 & \text{if } e > 0.9E
\end{cases}
\]
For these image datasets, we use two different CNN's as the \textit{read-in} networks; DenseNet \citep{densenet} for CIFAR10 and ResNet for FashionMNIST. The rationale for these models is motivated by the sentiment that including different CNN architectures with different image data gives more comprehensive support.

\subsubsection{FashionMNIST} \label{appendix:setup_fmnist_only}
For SPSA implementations, we require a perturbation coefficient $\mu=1e-2$, and SGD gives the best performance. These conclusions were reached after searching the following values [$1e-2$,$1e-3$,$1e-4$] as learning rates and delta values for both SGD and Adam.

{\leftskip=2em

\noindent Loss Function: Binary Cross Entropy.\newline
Optimizer: Adam ($\beta_1=0.9$, $\beta_2=0.999$, $\textit{weight\_decay}=0$ ), Epochs: 50. Batch size: 32. Seeds: [42, 63, 14]. GPU: A10G. \\

\subtitleblock{Read-in Network ($f_A$) - ResNet101}{
    Input/Output Dim: (28x28x1)/100. Activation: ReLU}
\subtitleblock{Black-Box Function ($\mathcal{R}$) - FFNN}{
    Input/Output Dim: 100/100. Hidden Layers: [500]. Activation: Tanh.}
\subtitleblock{Read-out Network ($f_B$) - FFNN}{
    Input/Output Dim: 100/10. Hidden Layers: [100,100]. Activation: Tanh.}
}

\subsubsection{CIFAR10}
For SPSA implementations, we require the perturbation coefficient $\mu=1e-3$, and Adam optimization is preferable over SGD. This was similarly determined over the grid search specified in \ref{appendix:setup_fmnist_only}.

{\leftskip=2em

\noindent Loss Function: Binary Cross Entropy.\newline
Optimizer: Adam ($\beta_1=0.9$, $\beta_2=0.999$, $\textit{weight\_decay}=0$ ), Epochs: 50. Batch size: 128. Seeds: [42, 63, 14]. GPU: A10G.

\subtitleblock{Read-in Network ($f_A$) - DenseNet}{
    Input/Output Dim: (32x32x3)/100. Activation: ReLU}
\subtitleblock{Black-Box Function ($\mathcal{R}$) - FFNN}{
    Input/Output Dim: 100/100. Hidden Layers: [500]. Activation: Tanh.}
\subtitleblock{Read-out Network ($f_B$) - FFNN}{
    Input/Output Dim: 100/10. Hidden Layers: [100,100]. Activation: Tanh.}
}

\subsection{Reservoir Scaling - CHP} \label{appendix:setup_calhouse_scaling}
\textsc{[Pretains to Figure~\ref{fig:reservoir_scaling}]}

In comparison to the setup of \ref{appendix:setup_sign_accuracy_and_convergence}, we increase the size of the networks and execute a grid search over the learning rates $1e-2$, $1e-3$, $5e-4$, $1e-4$ and $5e-5$. We found $1e-4$ to be the optimal initial learning rate for all methods except SPSA which required $1e-3$. We use the same lambda learning rate scheduler as \ref{appendix:setup_sign_accuracy_and_convergence}. For all experiments, we maintain an equal number of trainable parameters $|\theta_{A,B}| = 756,501$, aside from NFN(\textit{Rockmelon}) where $|\theta_{A,B}| = 616,601$ due to the bottleneck of the reservoir function.

Additional details include.

{\leftskip=2em

\noindent Loss Function: Huber.\newline
Optimizer: Adam ($\beta_1=0.9$, $\beta_2=0.999$, $\textit{weight\_decay}=0$), Epochs: 80. Batch size: 32. Seeds: [42, 63, 14]. GPU: A10G.

\subtitleblock{Read-in Network ($f_A$) - FFNN}{
    Input/Output Dim: 8/500. Hidden Layers: [500,500]. Activation: Tanh}
\subtitleblock{Black-Box Function [NFN(\textit{Fixed})] ($\mathcal{R}$) - FFNN}{
    Input/Output Dim: 500/500. Hidden Layers: [500]. Activation: Tanh.}
\subtitleblock{Black-Box Function [NFN(\textit{Parallel})] ($\mathcal{R}$) - FFNN}{
    Total Input/Output Dim: 500/500. Sub-reservoir Input/Output Dim: 5/5. Sub-Reservoir Hidden Layers: [100]. Number of Sub-Reservoirs: 100. Activation: Tanh.}
\subtitleblock{Black-Box Function [NFN(\textit{Rockmelon})] ($\mathcal{R}$) - Pre-Trained FFNN}{
    Total Input/Output Dim: 600/120. Sub-reservoir Input/Output Dim: 10/2. Sub-Reservoir Hidden Layers: [500,500,500,500]. Number of Sub-Reservoirs: 60. Activation: Tanh.}
\subtitleblock{Read-out Network ($f_B$) - FFNN}{
    Input/Output Dim: 500(or 120)/1. Hidden Layers: [500]. Activation: Tanh.}
}

\subsection{Reservoir Scaling - CIFAR100} \label{appendix:setup_cifar100}
\textsc{[Pretains to Figure~\ref{fig:cifar100}, and Tables~\ref{table:bondvsadam_sign},\ref{table:cifar100}]}

\begin{figure}[H]
    \centering
    \includegraphics[width=\textwidth]{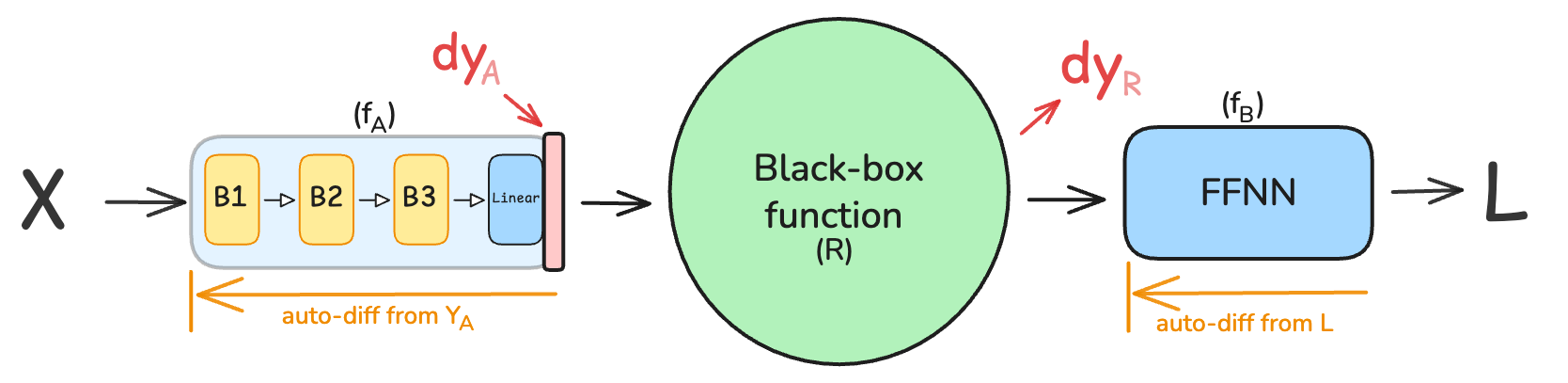}
    \caption{Schematic of NFN for CIFAR100 using DenseNet}
    \label{fig:NFN_CNN_diagram}
\end{figure}

For CIFAR-100 (as with CIFAR10), we use DenseNet \citep{densenet} to train the \textit{read-in} network. For clarity, we illustrate a schematic of the network structure in Figure~\ref{fig:NFN_CNN_diagram}. The CIFAR-100 training data incorporates standard augmentation techniques from torchvision's transforms library; including horizontal flipping, image cropping and rescaling, and normalization.

We use the same learning rate scheduler from setup~\ref{appendix:fmnist_cifar10_setup}, which deviates slightly from the scheduler set out on page 5 in \citet{densenet}, however we find that it obtains the best results in our implementation. \\

For CIFAR100 experiments, we use the following initial learning rates: ($\eta_{DenseNet}=1e-3$, $\eta_{FFNN} = 5e-4$) for BOND, BONDS and \enquote{no reservoir}, and ($1e-3$,$1e-6$) for SPSA. Additionally, SPSA uses SGD optimization whereas all other methods use Adam. These implementations were based off a grid search over ($1e-2$,$1e-2$), ($1e-3$,$1e-3$), ($5e-4$,$5e-4$), ($1e-4$,$1e-4$), ($1e-3$,$1e-5$) and ($1e-3$,$1e-6$) with both Adam and SGD.

Additional details.

{\leftskip=2em
\noindent Loss Function: Binary Cross Entropy. \newline 
Optimizer: Adam($\beta_1=0.9$, $\beta_2=0.999$, $\textit{weight\_decay}=0$). Epochs: 100. Batch size: 128. Seeds: [42, 63, 14]. GPU: A10G. \\
\subtitleblock{Read-in Network ($f_A$) - DenseNet}{
    Input/Output Dim: (32x32x3)/500(or 600). Activation: ReLU. Architecture: 3 DenseNet (Bottleneck) Blocks.}
\subtitleblock{Black-Box Function [NFN(\textit{Fixed})] ($\mathcal{R}$) - FFNN}{
    Input/Output Dim: 500/500. Hidden Layers: [500]. Activation: Tanh.}
\subtitleblock{Black-Box Function [NFN(\textit{Parallel})] ($\mathcal{R}$) - FFNN}{
    Total Input/Output Dim: 500/500. Sub-reservoir Input/Output Dim: 5/5. Sub-Reservoir Hidden Layers: [100]. Number of Sub-Reservoirs: 100. Activation: Tanh.}
\subtitleblock{Black-Box Function [NFN(\textit{Rockmelon})] ($\mathcal{R}$) - Pre-Trained FFNN}{
    Total Input/Output Dim: 600/120. Sub-reservoir Input/Output Dim: 10/2. Sub-Reservoir Hidden Layers: [500,500,500,500]. Number of Sub-Reservoirs: 100. Activation: Tanh.}
\subtitleblock{Read-out Network ($f_B$) - FFNN}{
    Input/Output Dim: 500(or 120)/100. Hidden Layers: [100,100]. Activation: Tanh.}
}

For clarity, we highlight that the augmented layer sizes to account for the input/output bottleneck of \textit{Rockmelon} actually reduces the overall trainable parameters. NFN(\textit{Rockmelon}) contains $3,785,756$ trainable and $4,543,758$ total parameters. NFN(\textit{Parallel}) contains $3,907,256$ trainable and $4,017,756$ total parameters. 

\newpage
\section{\textit{Rockmelon} Pre-Training} \label{appendix:setup_rockmelon}
\begin{figure}[H]
    \centering
    \includegraphics[width=0.5\textwidth]{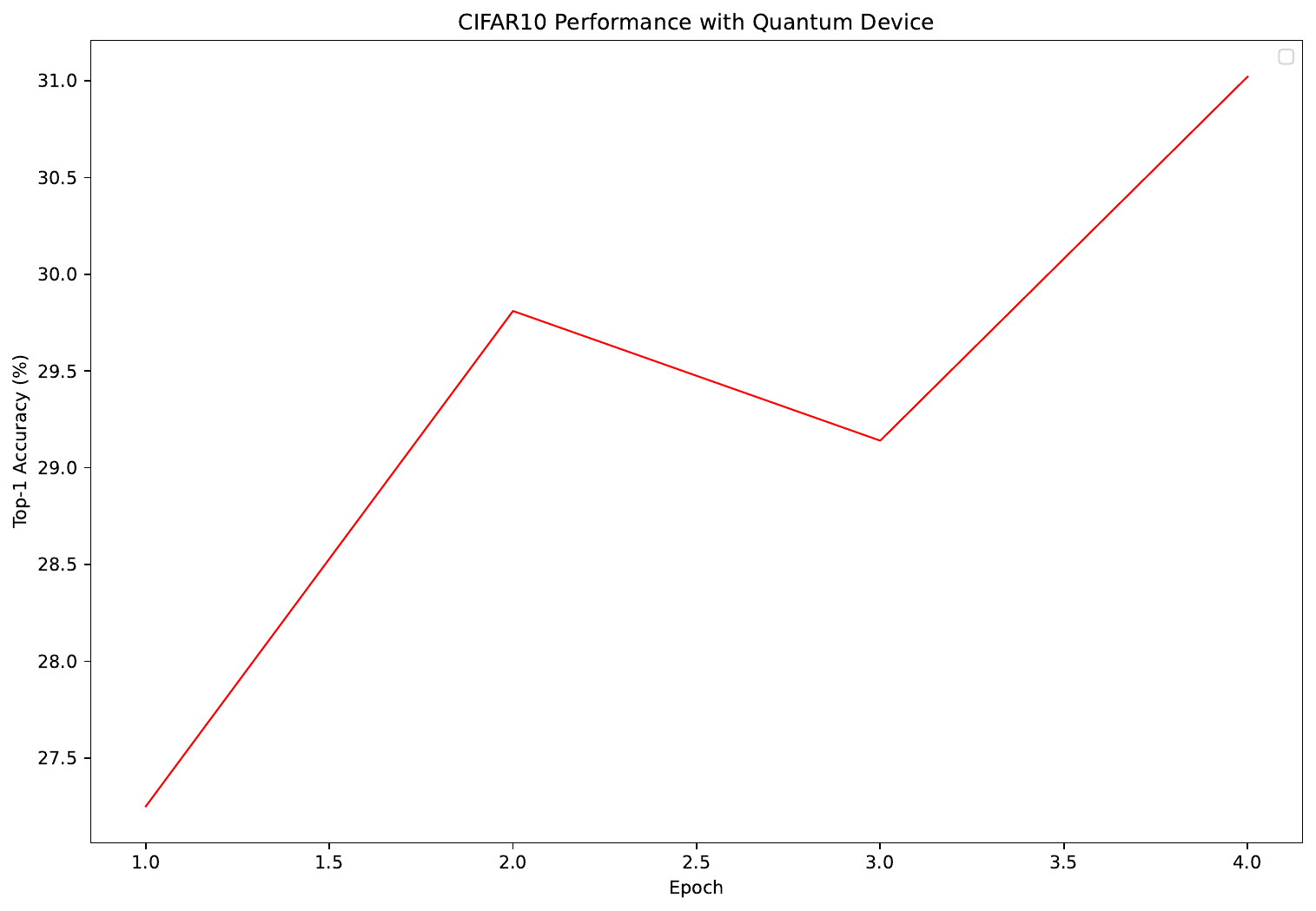}
    \caption{Performance of \textit{DenseNet - QR - FFNN} model on CIFAR10 using BOND}
    \label{fig:cifar10_qr}
\end{figure}

\begin{figure}[H]
    \centering
    \includegraphics[width=0.5\textwidth]{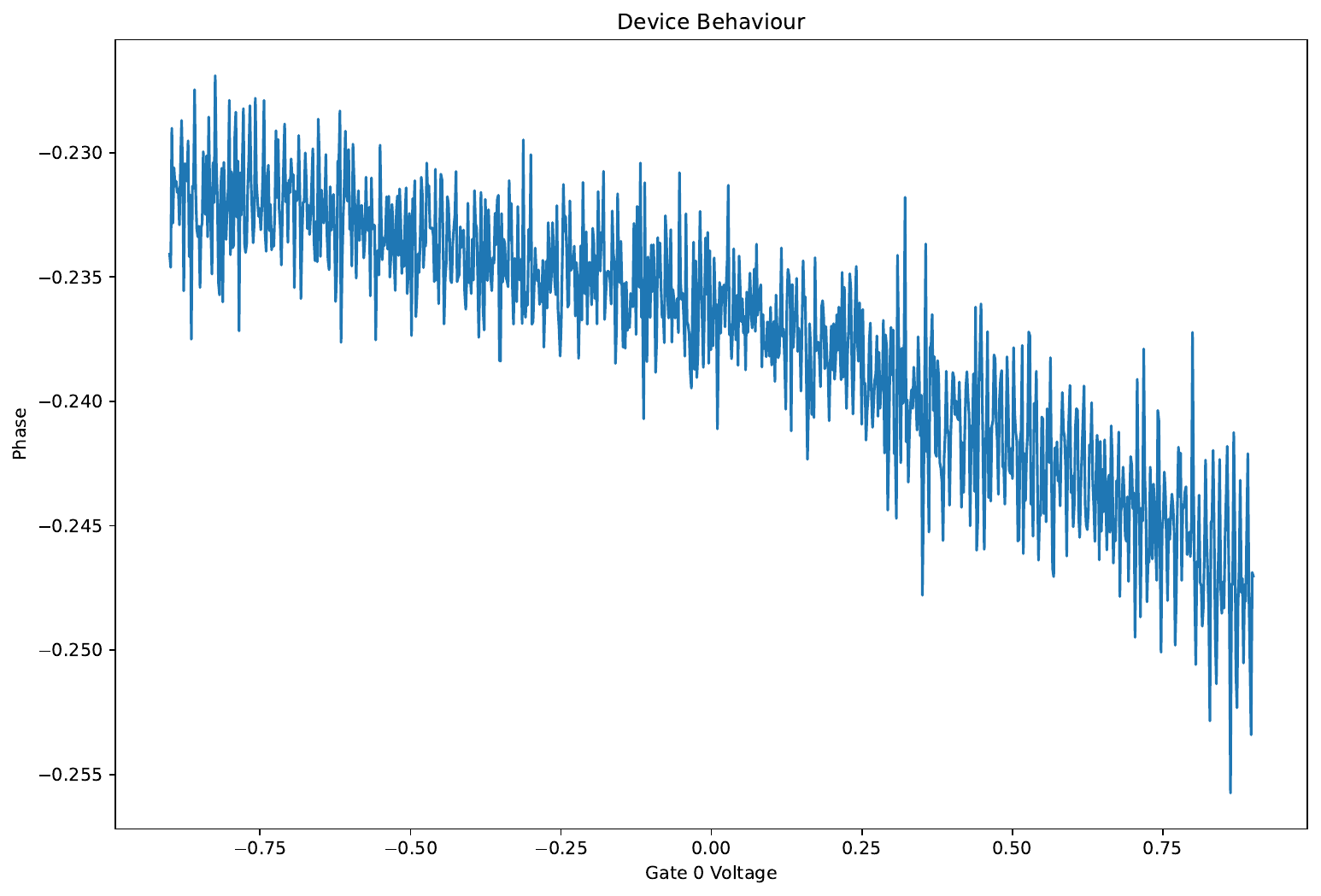}
    \caption{1-dimensional sweep of gate voltage on device}
    \label{fig:qr_sweep}
\end{figure}

\textit{Rockmelon} is an FFNN trained on input-output pairs sampled from a physical quantum RC device, the \enquote{Watermelon} quantum feature generator from Silicon Quantum Computing~\footnote{\href{https://sqc.com.au}{https://sqc.com.au}}. 
The device is a non-uniform array of phosphorus-doped silicon quantum dots, with 10 electrostatic control gates that determine electron trajectories when a source-drain bias is applied. Measuring current amplitude and phase at the drain provides the 2 device outputs. Integrating this device (in its current form) as a black-box module within a neural network proved challenging. The primary factors which affected performance and exploration capabilities are:
\begin{itemize}[noitemsep,topsep=0pt]
    \item Noise in the measurements
    \item Latency in converting tensors to physical voltages on the device, including ramping and resetting voltages between forward passes.
    \item Access to multiple devices, which limited the capacity for parallelization.
\end{itemize}
We highlight the CIFAR10 performance of a \textit{DenseNet - QR - FFNN} model in Figure~\ref{fig:cifar10_qr}, where \textit{QR} is a \textit{Parallel} configuration using a real quantum device as its sub-reservoir.
Due to latency constraints, we were not able to test beyond 4 epochs. For the findings of epochs 1-4, the performance is below what we would expect from a classical model. However, we suspect that the limited convergence is a function of noise in the measurement process on the physical device. Figure~\ref{fig:qr_sweep} shows a 1 dimensional sweep across the allowed input range of one gate.
As evidenced by Figure~\ref{fig:qr_sweep}, the non-smoothness of the device functionality impacts its performance and the capacity for gradient estimation when used directly. However, we consider that training a neural network on the input-output pairs of the device should create a smooth approximation of it. This leads to the modelling of the \textit{Rockmelon} reservoir, which we use as a staple throughout this work. Naturally \textit{Rockmelon} does not entirely capture the dynamics of the true device, but as an approximation, it encodes structured transforms which are not present in the random \textit{Parallel} and \textit{Fixed} reservoirs. The specific details for the pre-training of \textit{Rockmelon} are provided below.

{\leftskip=2em
\noindent Loss Function: Huber. \newline
Optimizer: Adam ($\beta_1=0.9$, $\beta_2=0.999$, $\textit{weight\_decay}=0$). Learning rate: $1e-4$. Epochs: 15. Batch size: 64. Input/Output Data: $1,000,000$ randomly sampled input-output pairs of the physical RC device.\\
\textbf{\textit{Rockmelon:}} Input/Output Dim: 10/2. Activation: Tanh. Hidden Layers: [500,500,500,500]
}

\noindent As the engineering of these devices continues to improve, factors such as noise and latency will likely be reduced, enabling further explorations into their utility/compatibility with reservoir computing implementations. We are excited for the opportunities that BOND may afford with these devices in future works.

\end{document}